\documentclass{article}
\pdfoutput=1  

\usepackage{cite} 
\usepackage{graphicx}
\usepackage{amssymb}

\newcommand{\eq}[1]{Eq.~\ref{eq.#1}} 
\newcommand{\eqbare}[1]{\ref{eq.#1}} 

\newcommand{\fig}[1]{Fig.~\ref{fig.#1}}

\newcommand{\tbl}[1]{Table~\ref{table.#1}}

\newcommand{\sect}[1]{Section~\ref{sect.#1}}

\newcommand{\sectlabel}[1]{\label{sect.#1}}
\newcommand{\eqlabel}[1]{\label{eq.#1}}
\newcommand{\figlabel}[1]{\label{fig.#1}}
\newcommand{\tbllabel}[1]{\label{table.#1}}

\newcommand{\figwidth}{3.5in}

\newcommand{\remove}[1]{}

\newcommand{\invivo}{\textit{in vivo}}
\newcommand{\exvivo}{\textit{ex vivo}}
\newcommand{\invitro}{\textit{in vitro}}

\newcommand{\BoltzmannConstant}{\ensuremath{k_B}}

\newcommand{\Tbody}{\ensuremath{T_{\mbox{\scriptsize body}}}}
\newcommand{\kThermal}{\ensuremath{k_{\mbox{\scriptsize thermal}}}}
\newcommand{\heatCapacity}{\ensuremath{c_{\mbox{\scriptsize thermal}}}}

\newcommand{\Pflux}{\ensuremath{P_{\mbox{\scriptsize flux}}}}
\newcommand{\Pradiated}{\ensuremath{P_{\mbox{\scriptsize flux}}^{\mbox{\scriptsize radiated}}}}
\newcommand{\Psignal}{\ensuremath{P_{\mbox{\scriptsize signal}}}}
\newcommand{\Pnoise}{\ensuremath{P_{\mbox{\scriptsize noise}}}}

\newcommand{\Router}{\ensuremath{R_{\mbox{\scriptsize outer}}}}
\newcommand{\Rinner}{\ensuremath{R_{\mbox{\scriptsize inner}}}}

\newcommand{\attenuationWater}{\alpha_{\mbox{\scriptsize water}}}
\newcommand{\attenuationLow}{\alpha_{\mbox{\scriptsize low}}}
\newcommand{\attenuationHigh}{\alpha_{\mbox{\scriptsize high}}}

\newcommand{\SNR}{{\mbox{SNR}}} 

\newcommand{\meter}{\mbox{m}}
\newcommand{\millimeter}{\mbox{mm}}
\newcommand{\micron}{\mbox{$\mu$m}}
\newcommand{\nanometer}{\mbox{nm}}
\newcommand{\second}{\mbox{s}}
\newcommand{\millisecond}{\mbox{ms}}
\newcommand{\microsecond}{\mbox{$\mu$s}}
\newcommand{\bit}{\mbox{bit}}
\newcommand{\GHz}{\mbox{GHz}}
\newcommand{\MHz}{\mbox{MHz}}
\newcommand{\kHz}{\mbox{kHz}}
\newcommand{\kg}{\mbox{kg}}

\newcommand{\picoWatt}{\mbox{pW}}
\newcommand{\Watt}{\mbox{W}}
\newcommand{\Joule}{\mbox{J}}
\newcommand{\zeptoJoule}{\mbox{zJ}}
\newcommand{\Kelvin}{\mbox{K}}
\newcommand{\Pascal}{\mbox{Pa}}

\newcommand{\bps}{\mbox{bits/s}}
\newcommand{\decibel}{\mbox{dB}}

\title{Acoustic Communication for Medical Nanorobots}
\author{Tad Hogg and Robert A.~Freitas Jr.\\Institute for Molecular Manufacturing\\Palo Alto,
CA}

\begin{document}

\maketitle

\begin{abstract}

Communication among microscopic robots (nanorobots) can coordinate their activities for biomedical tasks. The feasibility of \invivo\ ultrasonic communication is evaluated for micron-size robots broadcasting into various types of tissues. Frequencies between $10\,\MHz$ and $300\,\MHz$ give the best tradeoff between efficient acoustic generation and attenuation for communication over distances of about 100 microns. 
Based on these results, we find power available from ambient oxygen and glucose in the bloodstream can readily support communication rates of about $10^4 \bps$ between micron-sized robots. We discuss techniques, such as directional acoustic beams, that can increase this rate.
The acoustic pressure fields enabling this communication are unlikely to damage nearby tissue, and short bursts at considerably higher power could be of therapeutic use.

\end{abstract}

\section{Introduction}

Implanted or ingested medical devices can gather diagnostic
information and fine-tune treatments continually over an extended
period of time. Current examples include pill-sized
cameras to view the digestive tract as well as implanted glucose and
bone growth monitors to aid treatment of diabetes and joint
replacements, respectively.
The development of micromachines significantly extends the
capabilities of implanted devices. For example, clinical magnetic
resonance imaging (MRI) can move microrobots containing
ferromagnetic particles through blood
vessels~\cite{ishiyama02,\remove{mathieu05,}martel07a,\remove{yesin05}olamaei10}. Other
demonstrated micromachines use flagellar motors to move through
fluids, and offer the possibility of minimally invasive
microsurgeries in parts of the body beyond the reach of existing
catheter technology~\cite{behkam07,\remove{cole07,}fernandes09}.

Continuing the development of \invivo\ machines,
nanotechnology has the potential to revolutionize health
care~\cite{morris01,betancourt06\remove{nih03},thomas07,monroe09,sanchez09} with large numbers of devices, each of which is small
enough to reach and interact with individual cells of the body.
Current efforts focus on nanomaterials to enhance diagnostic imaging
and targeted drug delivery. For example, nanoparticles can target
specific cell types for imaging or drug
delivery~\cite{popovtzer08,sershen00,vodinh06,west00}. Other efforts
are developing more complex devices, such as multicomponent
nanodevices called tectodendrimers, which have a single core
dendrimer to which additional dendrimer modules of different types
are affixed, each type designed to perform a specific function~\cite{quintana00,baker01,betley02}.
These particles can also provide external control of some chemistry
within cells, such as through tiny radiofrequency (RF) antennas
attached to deoxyribonucleic acid (DNA) to control hybridization.

Further capabilities arise from combining the precision of these
nanoscale devices with the programmability currently only available
in larger machines. Such microscopic robots (``nanorobots''), with
size comparable to cells, could provide significant medical
benefits~\cite{freitas98,freitas99,freitas06,martel07,morris01,hill08}.
One approach to creating such robots is
engineering biological systems, e.g., RNA-based logic inside
cells~\cite{win08}, bacteria attached to
nanoparticles~\cite{martel08}, executing simple programs via the
genetic machinery within bacteria~\cite{ferber04,weiss06}, DNA
computers responding to specific combinations of
chemicals~\cite{benenson04} and artificial DNA-based robot-like structures capable of limited self-locomotion~\cite{smith10b}.
Another approach to manufacturing nanorobots is synthetic inorganic
machines~\cite{freitas10,freitas99}. Such fabrication is beyond current
technology, but could arise from
continued development of currently demonstrated nanoscale
electronics, sensors and
motors~\cite{barreiro08,berna05,collier99,craighead00,howard97,fritz00,marden02,montemagno99,wang05}
and relying on directed assembly~\cite{kufer08}, or from methods of molecular manufacturing involving mechanosynthesis~\cite{freitas08,freitas10a}.

In some tasks, robots could operate independently, e.g., each monitoring for specific chemical conditions~\cite{hogg06b} under which to release a drug~\cite{freitas99,freitas06} as an extension of an \invitro\ demonstration using DNA computers~\cite{benenson04}. More generally, robots could improve their performance by coordinating their actions, e.g., for nerve repair~\cite{hogg05}. In particular, communication over short distances could help robots avoid either too little or too large a response, improve sensing in tissue by using separate locations for transmission and detection of sensory signals, and form aggregated structures~\cite{freitas99}.  Bloodborne mobile nanorobots capable of short-range communication can share their onboard physiological data with other nanorobots that are stationary in the tissues, after which the stationary devices can pass the same data to other mobile nanorobots much later in time, thus enabling indirect long-distance messaging with no need for explicit long-range or external communication channels~\cite{freitas99}.  Likewise, fixed nanorobots can post messages to passing bloodborne devices which can then deliver those messages to fixed nanorobots stationed far from the site of origination, so either fixed or mobile nanorobots can serve a library function for \invivo\ robot systems.  These capabilities can enable complex nanorobot group behaviors, analogous to biological processes that recruit cells, e.g., for inflammation response -- but far better regulated, more flexible, and ultimately human-controlled.  Nanorobots with short-range communications can also establish navigational networks, including virtual maps such as vascular bifurcation detection~\cite{freitas99}.

Nanorobots could employ various communication methods~\cite{freitas99}. 
This paper quantitatively evaluates one such method, acoustic communication, that is well-suited for coordination over distances of around $100\,\micron$~\cite{freitas99}. We examine communication effectiveness, power requirements and effects on nearby tissue from the perspective of safety, sensing and therapeutic activities. We consider both isolated robots and groups of size up to about $10\,\micron$. Such groups include robots aggregated within vessels, giving larger effective radiator size and the ability to direct acoustic waves at useful frequencies.

We focus on the power requirements for acoustic communication between robots across various distances, in various directions, and through various tissues, and then present the results as a set of specifications for inter-robot acoustic communications. Specifically, the next section describes the relevant acoustic properties of tissues. We then consider two scenarios: an isolated robot and an aggregate of robots within a small blood vessel. The following sections describe directional beam capabilities, safety issues and several applications including communication rates in the presence of thermal noise. We conclude with recommendations for future work to extend this study.

\section{Nanorobot Acoustics in Tissue}

Acoustic behavior relevant for nanorobots depends on the physical properties of the robots and surrounding tissue and the operating frequency. This section summarizes the relevant properties and their numerical values.

\subsection{Acoustics}

Acoustics consists of small pressure variations in a fluid or solid medium. These variations satisfy the wave equation. We focus on the frequency response by taking the time dependence of acoustic quantities to oscillate with frequency $f$. The corresponding wavelength is $\lambda = c/f$, where $c$ is the speed of sound. 

The behavior of sound waves depends on both amplitude and phase of the waves, which are conveniently represented in combination as complex numbers. Specifically, we take the acoustic pressure at location $x$ and time $t$ to be $\Re(p(x) e^{-i \omega t})$ where $\omega = 2\pi f$ is the angular frequency of oscillation, $p(x)$ is a complex-valued amplitude for the pressure at the location, giving both the magnitude and phase of the pressure oscillation, and $\Re(\ldots)$ denotes the real part of the value. The magnitude $|p(x)|$ is the maximum variation in the pressure at location $x$ while the phase of $p(x)$ determines \emph{when} during an oscillation period the pressure reaches its maximum value. We  describe other quantities of interest, such as the velocity of the medium, with complex values, with the understanding that the physical quantity is the real part of the complex value multiplied by $e^{-i \omega t}$.

Using this choice of time dependence, the wave equation reduces to the Helmholtz equation~\cite{fetter80} 
\begin{equation}\eqlabel{wave equation}
\nabla^2 p(x) + k^2 p(x) = 0
\end{equation}
where $\nabla^2$ is the Laplacian differential operator and  $k$ is the complex-valued wave vector
\begin{equation}\eqlabel{wave vector}
k = \frac{\omega}{c} + i \alpha
\end{equation}
with $\alpha$ characterizing the attenuation of sound waves in the medium. 
For fluids, the attenuation is related to the dynamic and bulk viscosities, $\eta$ and $\xi$, respectively, by
\begin{equation}\eqlabel{viscosity}
\alpha = \left( \frac{4}{3}\eta + \xi \right) \frac{ \omega^2}{2 c^3 \rho }
\end{equation}
The corresponding velocity amplitude at location $x$ is
\begin{equation}\eqlabel{velocity}
v(x) = -\frac{i \omega }{c^2 k^2 \rho } \nabla p(x)
\end{equation}

\eq{wave equation} describes how sound propagates in the tissue. In our case, the source of the sound is oscillatory motion of nanorobot surfaces. Instead of considering internal structure of the nanorobot and how its mechanisms produce surface oscillations, we use boundary conditions specifying the velocity on the robot surface. We also require the amplitude of the sound to approach zero at large distances from the robot, i.e., we focus on the sound generated by the robot rather than other sources.
If only a part of the robot surface is actuated to produce oscillations, we assume the remainder of the robot surface does not move.

The robot must apply power to move its surface against the fluid. Due to the small robot sizes, we focus on Newtonian viscous effects as the most relevant~\cite{freitas99}, rather than non-Newtonian fluids or viscoelastic materials.
In a viscous fluid, pressure and viscosity produce forces acting on the robot surface. The $i^{th}$ component of the force the fluid exerts on a surface element $dA$ oriented in direction $\hat{n}$ is $-dA \sum_j T_{i,j} \hat{n}_j$, where the stress tensor component $T_{i,j}$ is the flux of the $i^{th}$ component of momentum density across a surface oriented with normal in direction $j$. For viscous fluids, the stress associated with small amplitude sound waves is
\begin{equation}\eqlabel{stress}
T_{i,j} = p \delta_{i,j} - \eta \left( \frac{\partial v_i}{\partial x_j} + \frac{\partial v_j}{\partial x_i}  \right) - \left(\xi - \frac{2}{3}\eta \right) \delta_{i,j} \nabla \cdot v
\end{equation}
where $ \delta_{i,j}=1$ if $i=j$ and is zero otherwise~\cite{fetter80}.
The force the robot must exert on the fluid to maintain the oscillation at its surface is the negative of the force from the fluid on the robot, namely
\begin{equation}\eqlabel{force}
F = \int_S T \cdot \hat{n} \,dA
\end{equation}
where the integral is over the oscillating surface $S$ of the robot. When pressure and velocity used to compute the stress in \eq{stress} are expressed in terms of the complex-valued amplitudes, the  time-dependent force is $\Re(F e^{-i \omega t})$ and the power applied by the robot on the fluid is 
\begin{equation}\eqlabel{power}
P(t) = \Re(F e^{-i \omega t}) \cdot \Re(v e^{-i \omega t})
\end{equation}
where $v$ is the velocity amplitude on the surface, assuming the robot imposes the same oscillation everywhere on the surface.
This expression gives the time-dependent power for the case of uniform velocity of the robot surface, which is the situation we focus on. In general, the robot could have different oscillation amplitudes -- both in magnitude and phase -- at different parts of the oscillating surface, in which case the power would be the integral of stress times velocity over the surface.

The power requirement can change sign during the oscillation period. That is, for some portion of the oscillation, the robot does work on the fluid while at other times the fluid does work on the robot. Provided the robot structure is elastic rather than dissipative, the robot could recover this power for subsequent use. Of more significance for quantifying communication power requirements is the time-averaged power, which in terms of the amplitudes is~\cite{fetter80}
\begin{equation}\eqlabel{average power}
P = \frac{1}{2} \Re(F \cdot v^*)
\end{equation}
where $v^*$ is the complex conjugate of $v$.

The radiated acoustic power arises from the pressure oscillations, with time-averaged power flux across a surface element $dA \hat{n}$ given by
\begin{equation}\eqlabel{average radiated power flux}
\Pradiated = \frac{1}{2} \Re(p \; v^* \cdot \hat{n})
\end{equation}
Integrating this flux over the surface gives the total power radiated through that surface.
We use \eq{average power} and \eq{average radiated power flux} to compute power dissipation by, and the acoustic radiated power of, the nanorobot.

\subsection{Acoustic Properties of Body Tissue}

\begin{table}[ht]
\centering
\begin{tabular}{ll}
parameter    &   value \\ \hline
speed of sound		& $c = 1500 \,\meter/\second$ \\
density			& $\rho = 1000 \,\kg/\meter^3$ \\
ambient temperature    &   $\Tbody=310\,\Kelvin$   \\
thermal conductivity    &   $\kThermal = 0.6\,\Watt/\meter/\Kelvin$   \\ 
heat capacity    &   $\heatCapacity = 4200\,\Joule/\kg/\Kelvin$   
\end{tabular}
\caption{\tbllabel{tissue parameters}Tissue parameters.}
\end{table}

Body tissues vary in acoustic properties, including speed of sound, density and attenuation. 
Boundaries between different types of tissue can scatter acoustic waves. For the relatively short distances we consider ($\sim 100\micron$, about 5 to 10 cell diameters) we consider a single type of tissue with homogeneous characteristics.
It is reasonable to ignore reflections from tissue boundaries at this scale because the amplitude of the reflection depends on the difference in acoustic impedance on either side of the boundary, which is very small because impedances for most tissues cluster narrowly between $1.4\mbox{--}1.8 \times 10^6 \,\kg/(\meter^2\,\second)$~\cite{freitas99}.  Scattering due to tissue inhomogeneities is also small because the wavelength at $100 \,\MHz$ in water is $15 \,\micron$, much larger than the typical $10\mbox{--}500 \,\nanometer$ size of intracellular organelles and other potential scattering foci in tissue.  
For speed of sound and density we use values corresponding to water, thus ignoring the small variations in these values over the relevant temperatures, frequencies and tissue types.
In particular, the speed of sound varies by less than 1\% in tissue over the range of frequencies we consider~\cite{akashi97,treeby09} and temperatures will be close to normal body temperature $\Tbody$. 

Attenuation varies considerably  with frequency. For pure fluids, \eq{viscosity} has attenuation increasing quadratically with frequency, i.e., increasing by a factor of 100 over the frequency range of $10\mbox{--}100\,\MHz$. At frequencies up to $10\,\MHz$ or so, biological tissue also shows a power-law increase in attenuation with frequency but with exponent mainly in the range $1$ to $1.5$. However, at the higher frequencies relevant for robot communication, this exponent gradually increases toward that of water.
\fig{attenuation} shows measured attenuation in some biological tissues and the fits we use for three cases. Specifically, the attenuation curves for water~\cite{akashi97,daft89} and the two types of tissue are
\begin{eqnarray}
\attenuationWater(f) &=& 0.025 f^2 \nonumber \\
\attenuationLow(f) &=& 0.5 f^{1.36} + 0.025 f^2   \eqlabel{attenuation} \\
\attenuationHigh(f) &=& 5.2 f^{1.28} + 0.137 f^2 \nonumber
\end{eqnarray}
with attenuation measured in $\meter^{-1}$ and frequency in $\MHz$.

\begin{figure}[t]
\centering \includegraphics[width=5in]{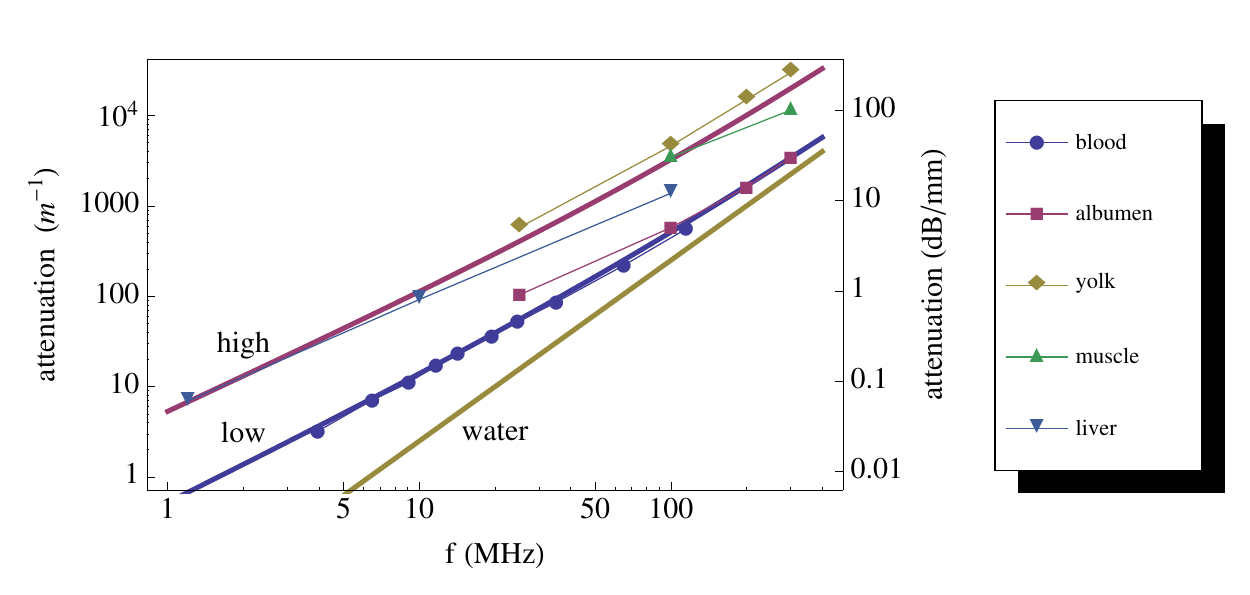} 
\caption{Attenuation as a function of frequency ($f$). Attenuation values are measured in $\meter^{-1}$ on the left and $\decibel/\mbox{mm}$ on the right. The thick curves show the fits we use for high and low tissue attenuation, and for water~\cite{akashi97,daft89}. The points are measured values in blood~\cite{goss80}, albumen and egg yolk~\cite{akashi97}, and muscle and liver~\cite{daft89}.}\figlabel{attenuation}
\end{figure}

\begin{table}[t]
\begin{center}
\begin{tabular}{lcc}
frequency				& $10\,\MHz$ 	& $100\,\MHz$ \\
wavelength			& $150\,\micron$	& $15\,\micron$ \\
 & \multicolumn{2}{c}{attenuation distance} \\
water		& $4\times10^5\,\micron$	& $4000\,\micron$ \\
low			& $7\times10^4\,\micron$	& $2000\,\micron$ \\
high			& $9\times10^3\,\micron$	& $300\,\micron$ \\
\end{tabular}
\end{center}
\caption{\tbllabel{sound parameters}Acoustic parameters, including the characteristic attenuation distance, $1/\alpha(f)$, for the three cases we consider.}
\end{table}

We mainly focus on two frequencies, shown in \tbl{sound parameters}, covering the range giving a good compromise between acoustic radiation efficiency (favoring higher frequencies) and attenuation (favoring lower frequencies). For communication over distances of about $100\,\micron$, decrease in power due to attenuation, given by $\exp(-2 \alpha d)$ for distances $d$ large compared to the size of the acoustic source, is relatively small. Instead, the decreasing power flux due to the spread of the acoustic wave (proportional to $1/d^2$) is the significant limitation on communication distance and attenuation affects efficiency mainly through its relation to viscous dissipation at the emitting robot surface via \eq{viscosity}.

\section{Spherical Nanorobots}
\sectlabel{sphere}

\begin{figure}[ht]
\centering \includegraphics[width=\figwidth]{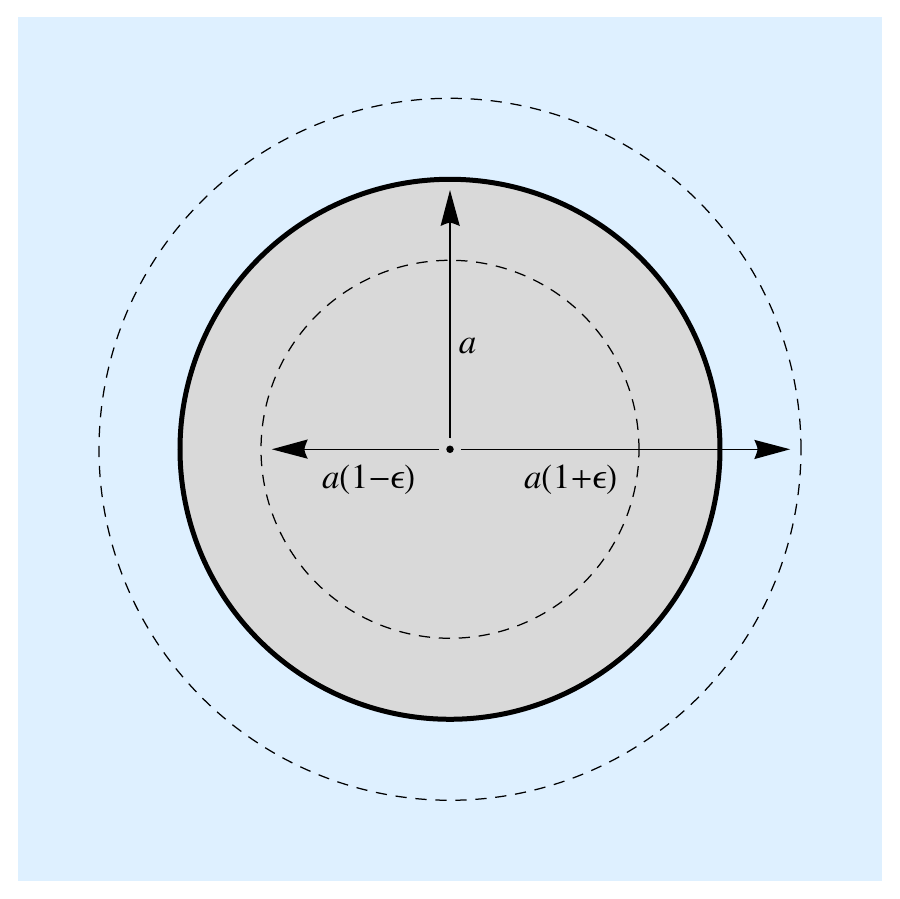}
\caption{Radial pulsations of a sphere, ranging a distance $\epsilon a$ on either side of the undisturbed radius $a$. For sound waves, $\epsilon \ll 1$, so the range of oscillation shown in this figure is greatly exaggerated. The cases we consider have $\epsilon$ ranging from $10^{-9}$ to $10^{-3}$.}\figlabel{sphere geometry}
\end{figure}

Nanorobots could have a variety of shapes, depending on the task and manufacturing constraints. One example is spherically shaped robots for use within blood vessels~\cite{freitas98}. A sphere is a particularly simple geometry for studying nanorobot capabilities. Thus, as a model of acoustic communication for an isolated robot we consider the pulsating sphere shown in \fig{sphere geometry}. The sphere's radius changes as $a+a \epsilon \cos(\omega t)$ \remove{or $a + \Re(a \epsilon e^{-i \omega t})$ }with $\epsilon \ll 1$.

\subsection{Sound Pressure and Velocity}

Due to the spherical symmetry of this geometry, the acoustic pressure and velocity depend only on the distance from the sphere, and velocity is directed radially. Thus acoustic radiation is the same in all directions so \eq{wave equation} becomes
\begin{equation}
\frac{d^2 p}{d r^2} + \frac{2}{r} \frac{d p}{d r} + k^2 p = 0
\end{equation}
The solution matching the motion of the sphere at $r=a$ and decreasing to zero as $r \rightarrow \infty$ is
\begin{equation}\eqlabel{sphere pressure}
p(r) = \frac{a^3 c^2 k^2 \rho  \epsilon}{r (-1+i a k)} \;  e^{-i k(a-r)}
\end{equation}
The corresponding velocity, from \eq{velocity}, is
\begin{equation}\eqlabel{sphere velocity}
v(r) = -\frac{i a^3 \omega  \epsilon  (k r+i) }{r^2 (a k+i)} \;  e^{-i k (a-r)}
\end{equation}
The velocity at the surface of the sphere is $v(a) =  -i a \omega  \epsilon$. Thus the time-dependent velocity $\Re(v(a) e^{-i \omega t}) = -a \omega  \epsilon  \sin (\omega t)$ corresponds to the motion of the sphere specified above.

The wave vector $k$, given in \eq{wave vector}, has positive imaginary part so the last factor in these expressions
\begin{displaymath}
e^{-i k (a-r)} = e^{-\alpha (r-a)}  e^{i \omega (r-a)/c}
\end{displaymath}
decreases exponentially as $r \rightarrow \infty$.
Since the attenuation distances (\tbl{sound parameters}) are, in most cases, large compared to the robot sizes and communication distances we consider, the exponential attenuation factor $e^{-\alpha (r-a)}  \approx 1$ for these distances. That is, the attenuation factor is a relatively minor contribution to acoustic power loss. Over these distances, \eq{sphere pressure} shows the pressure decreases as $1/r$. The velocity behavior depends on the frequency. If $k a$ is small (low frequency or a small sphere), then $v(r)$ decreases rapidly, as $1/r^2$ near the sphere, up to distances where $k r \approx 1$, beyond which the velocity decreases more slowly, as $1/r$. At high frequencies or for larger spheres, when $k a$ is comparable or larger than one, the velocity decreases at the slower rate $1/r$ over this full range of distance.
Thus the value $p \; v^*$ determining the radiated power flux (\eq{average radiated power flux}) decreases as $1/r^3$ for low frequencies and near the sphere or as $1/r^2$ otherwise. While the relative phases of $p$ and $v$ also determine the time-average radiated power, this behavior for the magnitude of $p \; v^*$ indicates situations with $k a \ll 1$, i.e., a small sphere or using low frequencies, are less effective at radiating acoustic power. In tissue, this effect is compounded by the higher tissue viscosity at low frequencies.
Due to the spherical symmetry, the total radiated power is the flux times the surface area $4 \pi r^2$. Thus power flux decreasing as $1/r^2$ corresponds to constant total radiated power, the highest possible efficiency.

\subsection{Transmission Capability}

We consider three sizes for the robots to illustrate acoustic transmission capabilities. First, a radius of $0.5\,\micron$ corresponds to a typical individual isolated nanorobot small enough to passively move through the circulatory system with the blood flow. Second, a $5\,\micron$ radius is roughly the size of the ringset aggregates considered in \sect{ringset} and is about the upper size limit for robots that could move actively through the circulation~\cite{freitas07}. 
And finally, a $50\,\micron$ radius would be the size of tissue-embedded repeater stations for a communication network. The larger robots could be placed directly into the tissue via microneedles or form via self-assembly~\cite{freitas09} from smaller robots that have diapedesed out of nearby blood vessels.

\fig{sphere pressure 10} shows the behavior of sound around a pulsating sphere at $10\,\MHz$. The plots show both the variation in power used by the sphere over a single oscillation period and the sound field at the four times during that period with the extreme values of power. The maximum power consumption occurs at times when the sphere is contracting away from low pressure or expanding into high pressure. At the power minima, the sphere is contracting away from high pressure or expanding into low pressure, in which case the fluid does work on the sphere.
At this frequency, the variation in pressure and velocity are nearly $90^\circ$ out of phase: the largest pressures (both positive and negative) occur when the speed of the sphere is close to zero.
As seen in \fig{sphere pressure 100}, at higher frequencies more of the motion of the sphere goes into generating outgoing sound waves. This behavior is a consequence of the variation in pressure and velocity at the sphere surface being nearly in phase.

\begin{figure}[thp]
\begin{center}
\begin{tabular}{c}
\includegraphics[width=\figwidth]{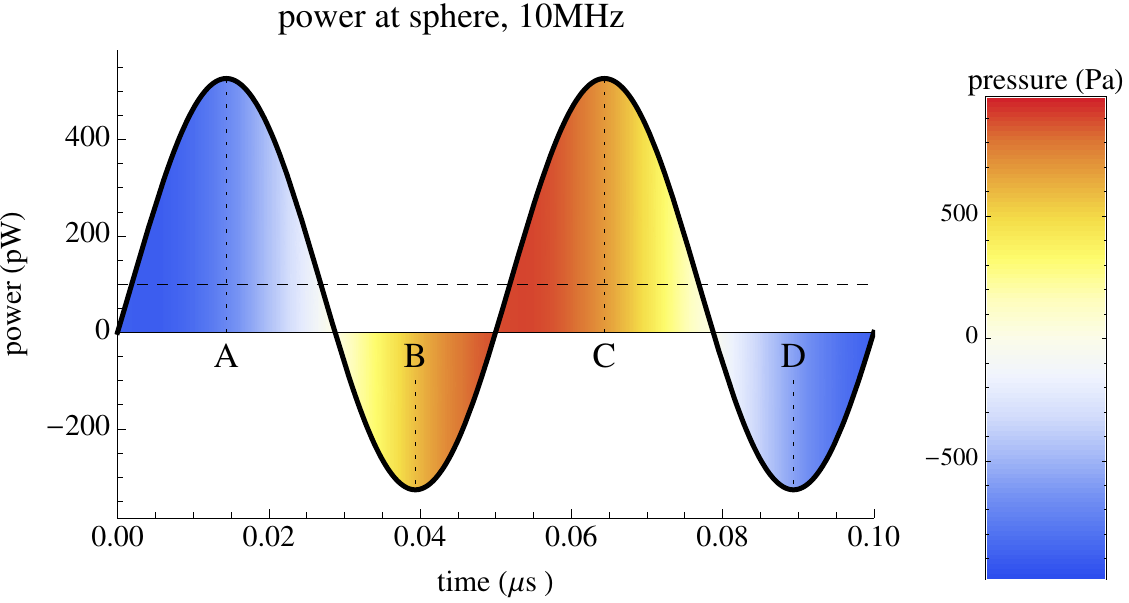} \\
\includegraphics[width=\figwidth]{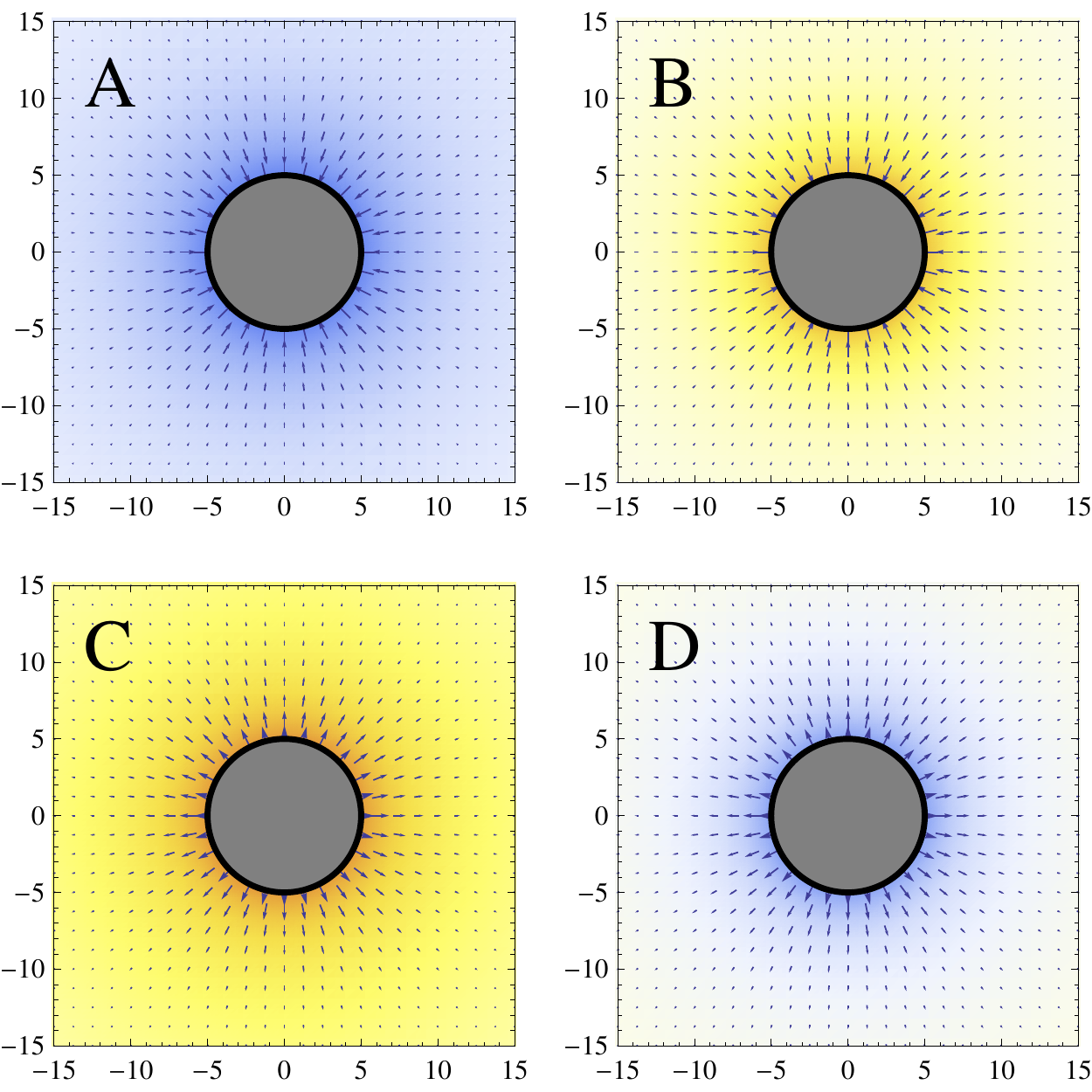} \\
\end{tabular}
\end{center}
\caption{Sound from a  $5\,\micron$ radius sphere at $f=10\,\MHz$ in low-attenuation tissue. (Top) Power expended by the sphere against the fluid as a function of time during one oscillation period for motion with time-averaged power of $100\,\picoWatt$ (indicated by the dashed horizontal line). Colors indicate the pressure at the surface of the sphere at each time. Letters indicate times of successive maxima and minima in power. (Bottom) Pressure and velocity at four times during an oscillation period, corresponding to successive maxima and minima in power as indicated by the letters in each corner. Plots on the left are at the two times of maximum power use. Arrows at the surface of the sphere show velocity of the sphere and fluid immediately next to the sphere, other arrows show the fluid velocity at various distances from the surface.
Distances along both axes are in microns and the gray disk is the location of the sphere.}\figlabel{sphere pressure 10}
\end{figure}

\begin{figure}[thp]
\begin{center}
\begin{tabular}{c}
\includegraphics[width=\figwidth]{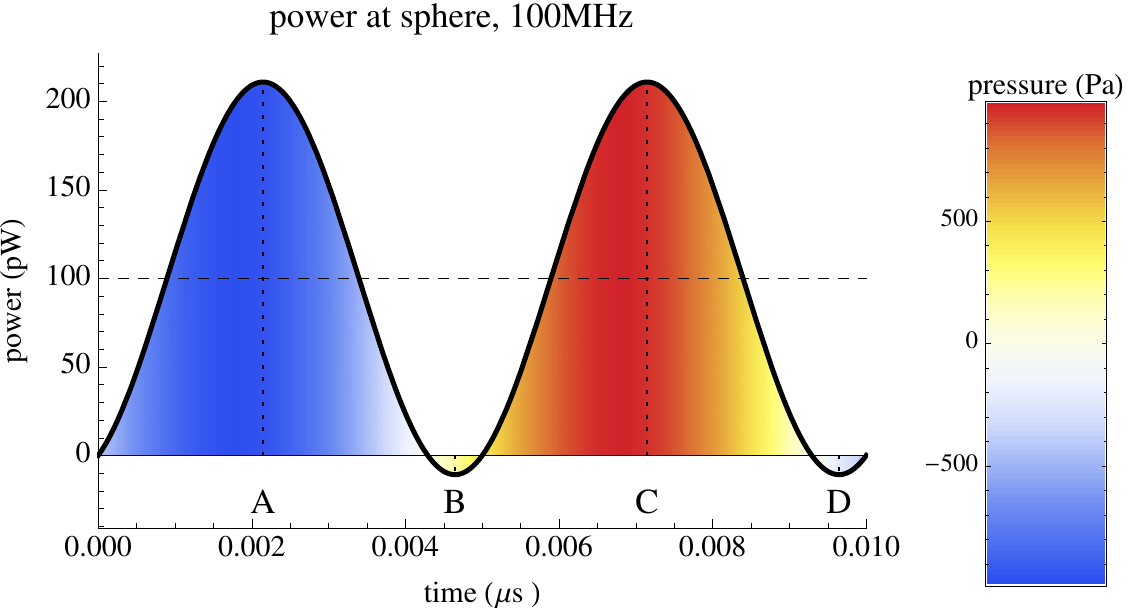} \\
\includegraphics[width=\figwidth]{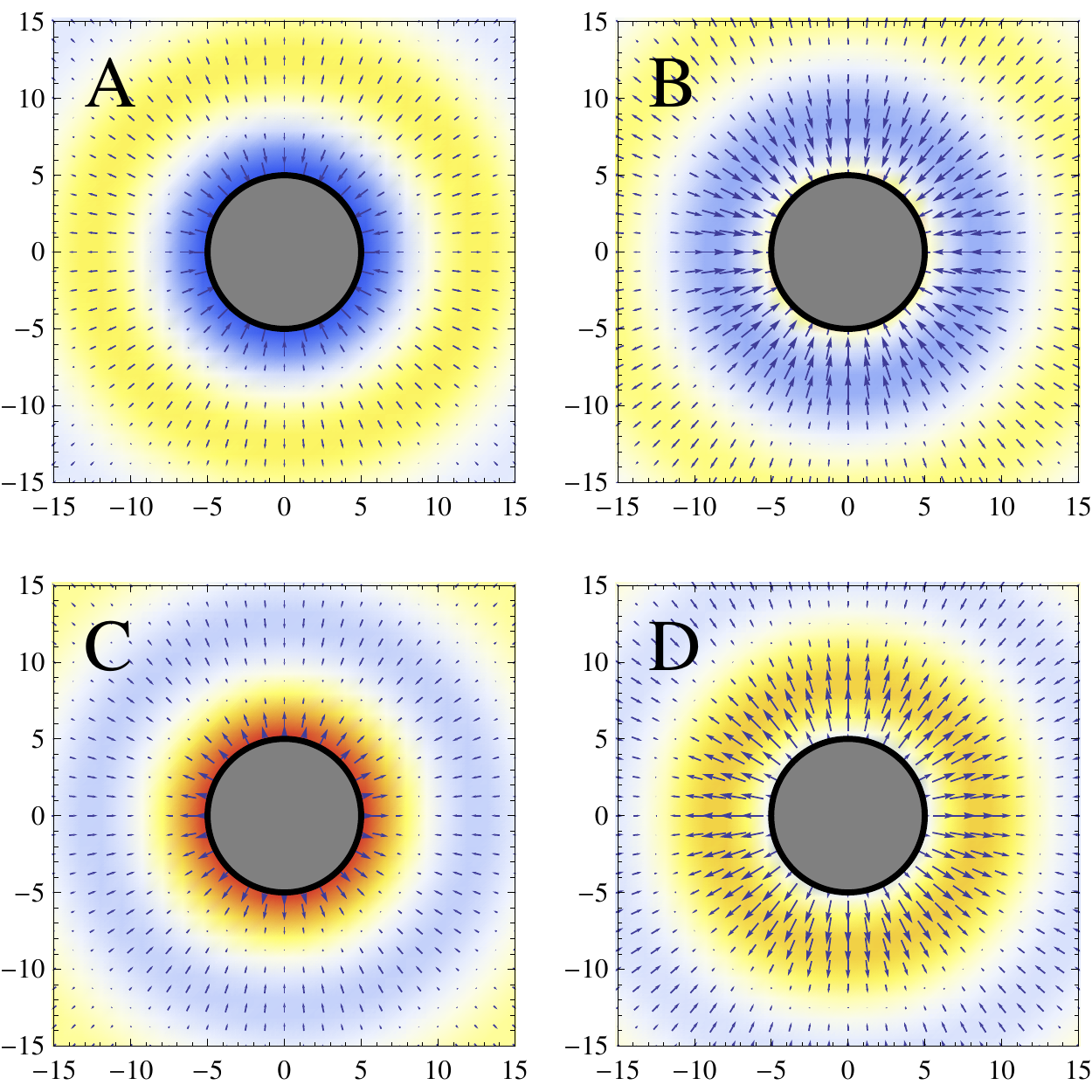} \\
\end{tabular}
\end{center}
\caption{Sound from a  $5\,\micron$ radius sphere at $f=100\,\MHz$ in low-attenuation tissue. (Top) Power expended by the sphere against the fluid as a function of time during one oscillation period for motion with time-averaged power of $100\,\picoWatt$ (indicated by the dashed horizontal line). Colors indicate the pressure at the surface of the sphere at each time, using the same range as \fig{sphere pressure 10}. Letters indicate times of successive maxima and minima in power. (Bottom) Pressure and velocity at four times during an oscillation period, corresponding to successive maxima and minima in power as indicated by the letters in each corner. Plots on the left are at the two times of maximum power use. Arrows at the surface of the sphere show velocity of the sphere and fluid immediately next to the sphere, other arrows show the fluid velocity at various distances from the surface.
Distances along both axes are in microns and the gray disk is the location of the sphere.}\figlabel{sphere pressure 100}
\end{figure}

We pick the oscillation amplitude so that the time-average power (\eq{average power}) is $100\,\picoWatt$. If all this power is radiated without attenuation, the power flux at $100\,\micron$ would be $8 \times 10^{-4}\,\picoWatt/\micron^2$.
\tbl{sphere results} shows the behavior for spheres of different sizes. For the small sphere, most of the input power is dissipated by viscous forces, especially at the lower frequency where tissue viscosity is large. Power efficiency for the larger sphere is close to 100\% for both frequencies, and maximum pressure variation is fairly small.

\begin{table}[tb]
\begin{center}
\begin{tabular}{lcc}
frequency	 ($\MHz$)	& 10	& 100 \\
\hline
\multicolumn{3}{c}{radius $0.5\,\micron$} \\
radiated power at $100\,\micron$ ($\picoWatt$) & $0.7$	&  $58$\\
average power flux at $100\,\micron$ ($\picoWatt/\micron^2$) & $5.4\times 10^{-6}$	& $4.6\times 10^{-4}$ \\
maximum pressure (\Pascal) 	& $800$	& $8000$	\\
\hline \multicolumn{3}{c}{radius $5\,\micron$} \\
radiated power at $100\,\micron$ ($\picoWatt$) & $87$	&  $90$\\
average power flux at $100\,\micron$ ($\picoWatt/\micron^2$) & $6.9\times 10^{-4}$	& $7.2\times 10^{-4}$ \\
maximum pressure (\Pascal) 	& $910$	& $980$	\\
\hline \multicolumn{3}{c}{radius $50\,\micron$} \\
radiated power at $100\,\micron$ ($\picoWatt$) & $100$	&  $95$\\
average power flux at $100\,\micron$ ($\picoWatt/\micron^2$) & $7.9\times 10^{-4}$	& $7.6\times 10^{-4}$ \\
maximum pressure (\Pascal) 	& $98$	& $98$	\\
\end{tabular}
\end{center}
\caption{\tbllabel{sphere results}Behavior of a pulsating sphere of various sizes in tissue with low attenuation.}
\end{table}

\begin{figure}[htbp]
\centering 
\includegraphics[width=6in]{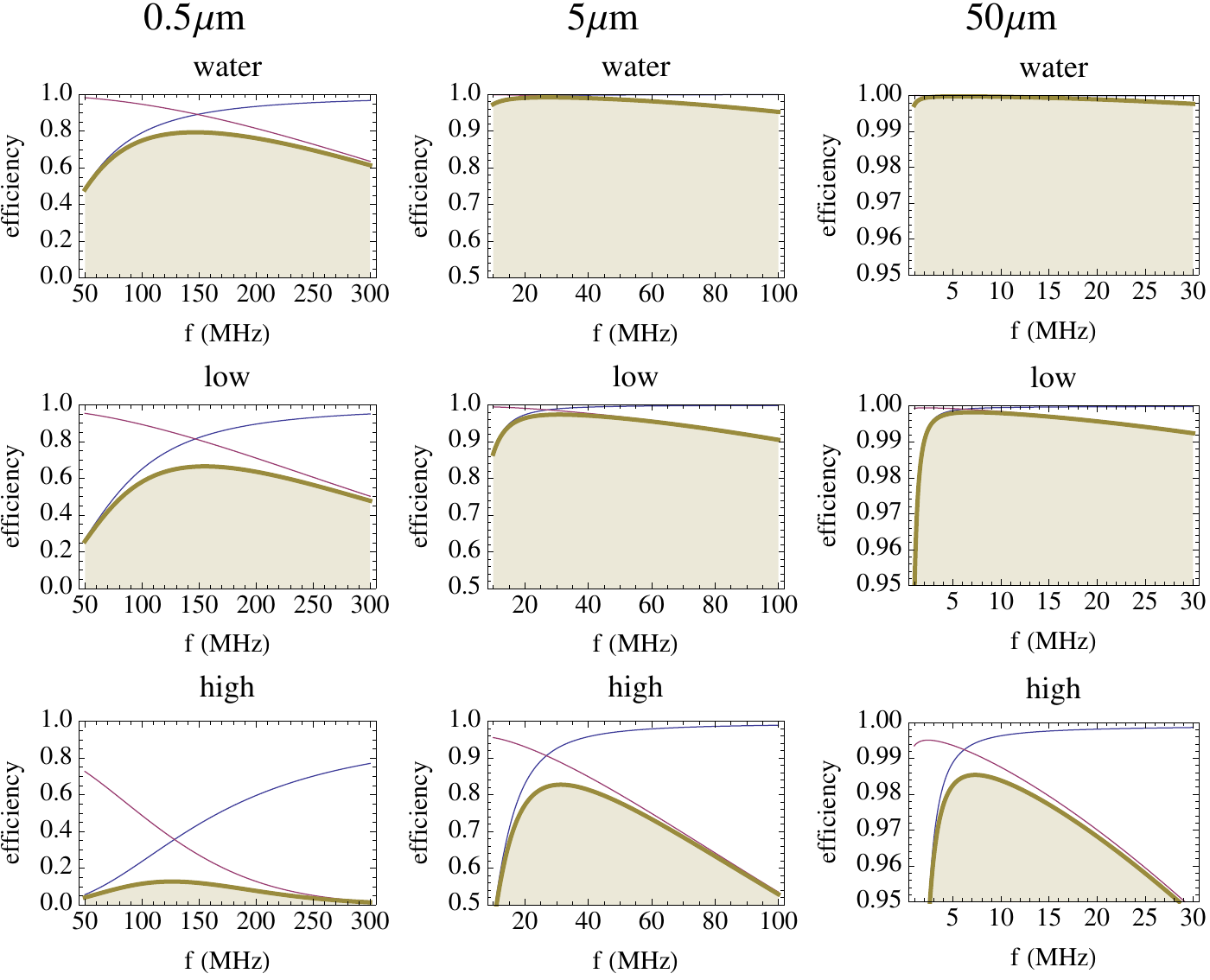}
\caption{Contributions to broadcast power efficiency at $100\micron$ from a pulsating sphere at various frequencies, sphere sizes and tissue attenuations. Each column shows behavior for a sphere with the radius indicated at the top of that column. The attenuations are given by \eq{attenuation}, with each row corresponding to one of the attenuation cases.
The thin curves in each plot show the contributing factors to the efficiency: the acoustic efficiency (i.e., fraction of input power producing acoustic waves at the sphere surface) which increases with frequency and the transmission efficiency at $100\,\micron$ (i.e., fraction of acoustic energy produced by the sphere that reaches $100\,\micron$) which decreases with frequency. The overall efficiency (thick curve) is the product of these two factors.
The vertical and horizontal axes have different ranges for the different sphere sizes.
}\figlabel{sphere efficiency}
\end{figure}

\fig{sphere efficiency} shows how efficiency results from a combination of losses due to viscous forces at the sphere and attenuation at a distance of $100\,\micron$ as determined with \eq{sphere pressure} and \eqbare{sphere velocity}. Specifically, overall efficiency for transmitting acoustic signals over a distance $d$ is defined by the ratio of the time-averaged acoustic power radiated to distance $d$ to the time-averaged power required to make the sphere surface oscillate. This ratio is the product of two factors. The first factor is the \emph{acoustic efficiency}, i.e., the fraction of the robot's power that produces outgoing acoustic radiation rather than dissipated against viscous forces. The second factor is the \emph{transmission efficiency}, i.e., the fraction of outgoing power generated at the sphere that reaches distance $d$ rather than being lost to attenuation in the tissue. Acoustic efficiency and attenuation both increase with frequency, giving a tradeoff in selecting the best frequency for communication. \fig{sphere efficiency} shows that frequencies between $10$ and $150\MHz$ provide a good tradeoff for communication among microscopic robots. 
The peaks in efficiency for the tissue attenuations we consider are around $150 \,\MHz$ for $0.5\,\micron$ robots, $30\,\MHz$ for $5\,\micron$ robots, and $5\,\MHz$ for $50\,\micron$  robots.

While these results indicate operating frequencies that optimize power efficiency, the exact values depend on the physical parameters such as attenuation that are not precisely known. Thus the main point of these results is the existence of the tradeoff and the range of frequencies giving the best performance. Precise operating choices will require knowledge of the exact physical properties of the specific tissues nanorobots will operate in. Such robots will be useful research tools for determining these properties, thereby allowing calibration of robot operation for individual tissues.

Moreover, the maximum data rate is an important criterion in addition to power efficiency. Since data rate increases with frequency, it may be useful to operate at somewhat higher frequencies than those providing the best power efficiency. For example, a $5\,\micron$ robot has efficiency decreasing from $0.97$ at $30\,\MHz$ to $0.90$ at $100\,\MHz$ in a low-attenuation environment.

\section{Nanorobot Ringsets on a Vessel Wall}
\sectlabel{ringset}

\begin{figure}[htbp]
\begin{center}
\begin{tabular}{lc}
\raisebox{4.5cm}{(a)} & \includegraphics[width=\figwidth]{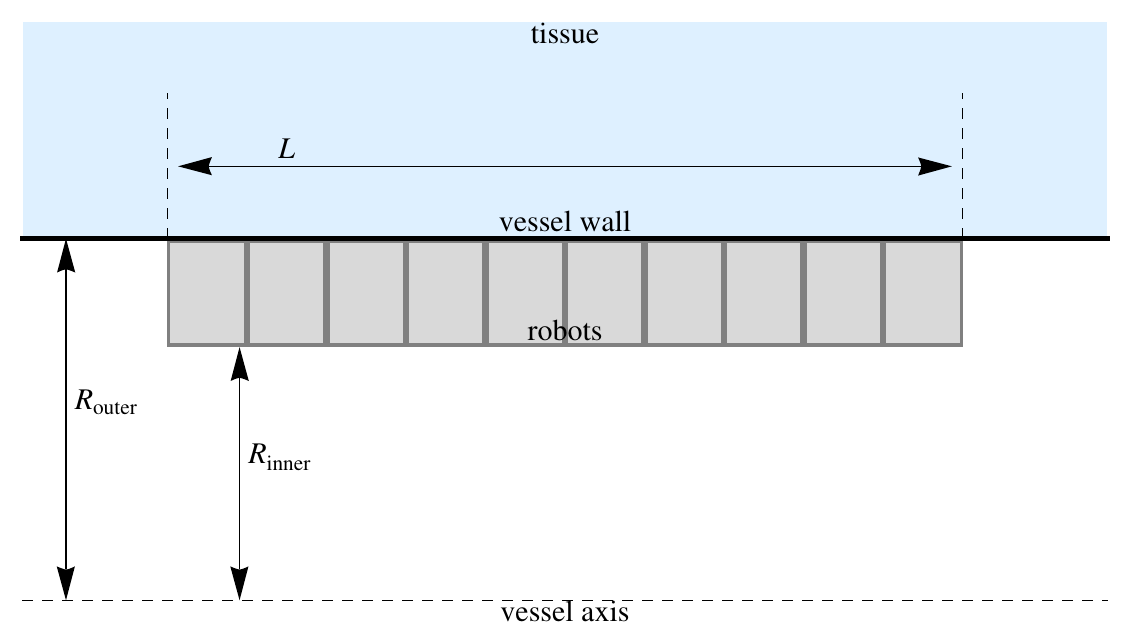}\\
& \\
\raisebox{8cm}{(b)} & \includegraphics[width=\figwidth]{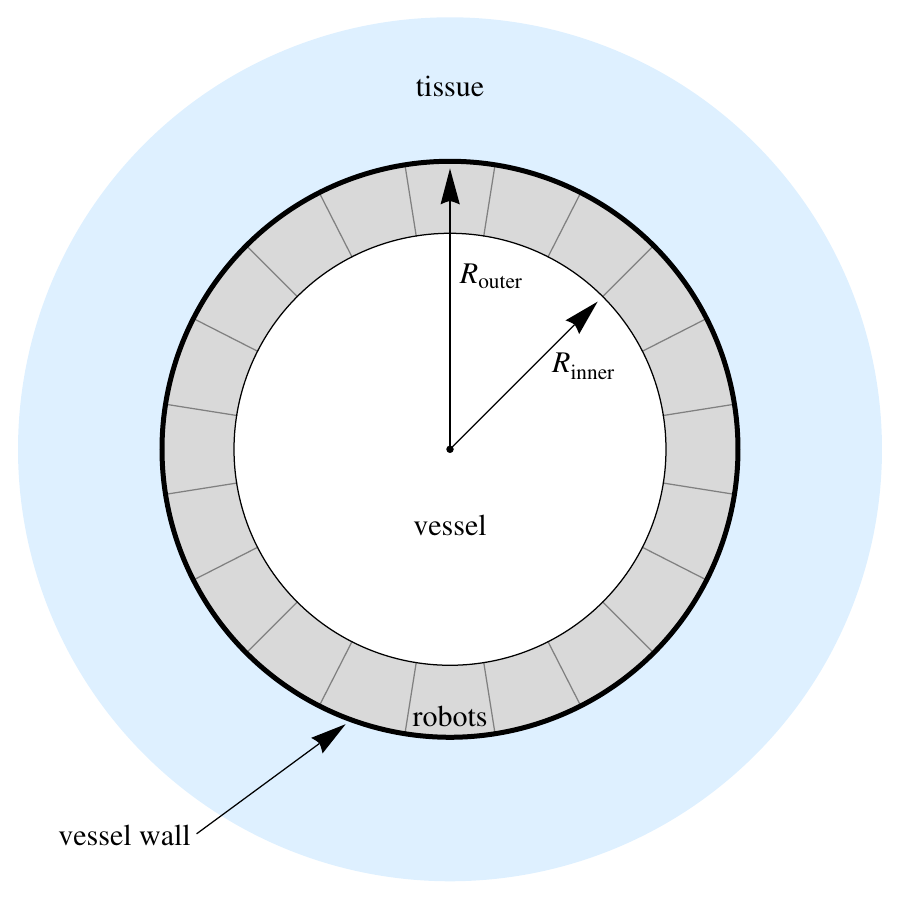} \\
\raisebox{3cm}{(c)} & \includegraphics[width=\figwidth]{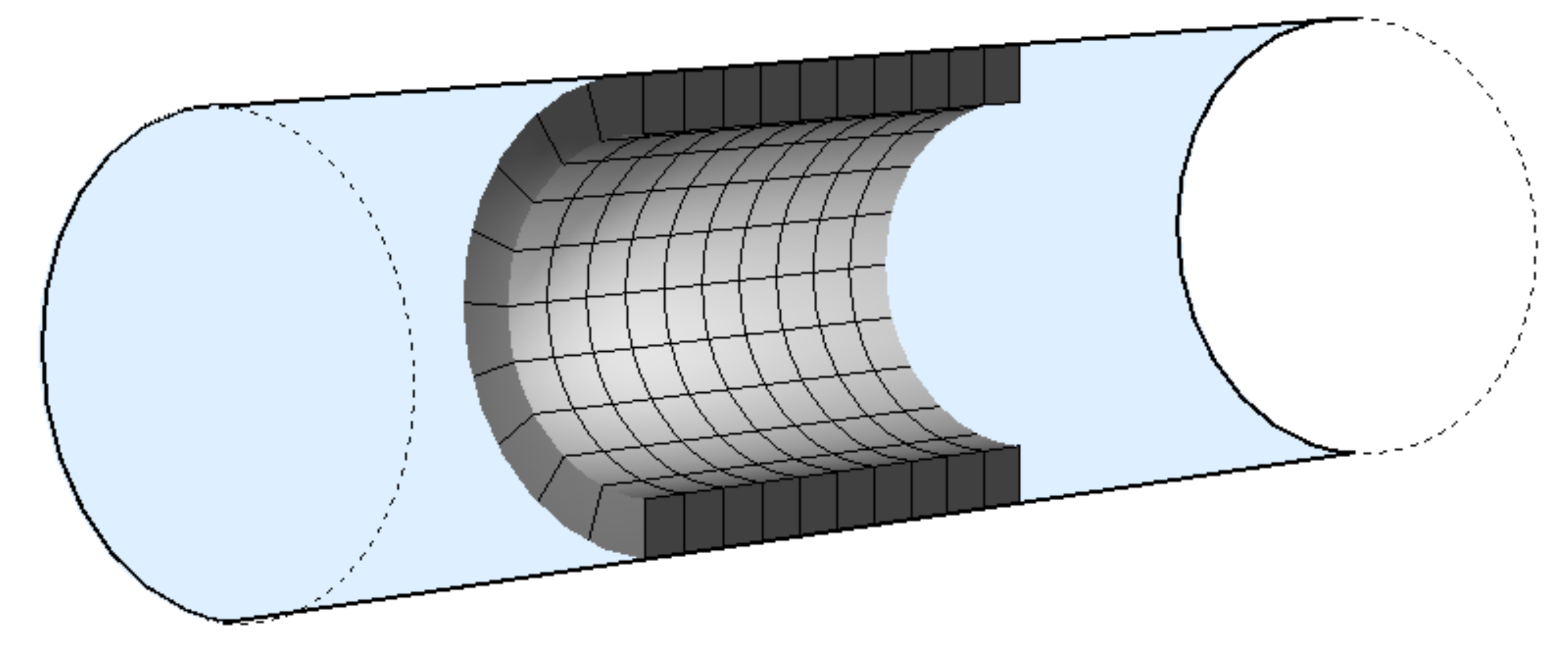} \\
\end{tabular}
\end{center}
\caption{Robot ringset inside a small vessel: (a) longitudinal, (b) cross section and (c) 3D views. The 3D view shows half the vessel.}\figlabel{ringset geometry}
\end{figure}

\begin{figure}[ht]
\centering \includegraphics[width=\figwidth]{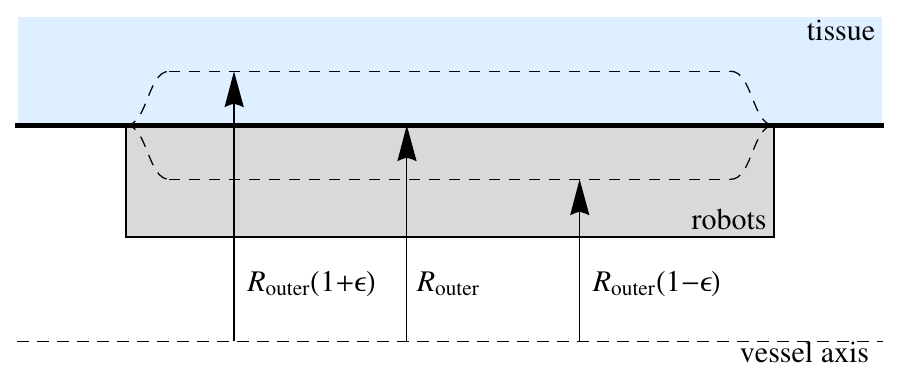} 
\caption{Range of oscillation of outer surface of robot ringset, ranging a distance $\epsilon \Router$ on either side of the undisturbed size $\Router$. To avoid discontinuity in the surface, the range of motion smoothly decreases to zero at either end of the ringset, as illustrated. For sound waves, $\epsilon \ll 1$, so the range of oscillation shown in this figure is greatly exaggerated.  The cases we consider have $\epsilon$ ranging from $10^{-7}$ to $10^{-5}$.}\figlabel{ringset movement}
\end{figure}

Individual robots moving passively with the circulation can approach within a few cell diameters of most tissue cells of the body. To enable passing through even the smallest vessels, the robots must be at most a few microns in diameter. This small size limits the capabilities of individual robots. For tasks requiring greater capabilities, robots could form aggregates via self-assembly~\cite{freitas09}. For robots reaching tissues through the circulation, the simplest aggregates are formed on the inner wall of the vessel, forming circumferential rings of robots that are then organized into still larger aggregates of adjacent rings, called ``ringsets"~\cite{hogg10}.  Nanorobot ringsets positioned in one location on luminal surfaces of capillaries or other blood vessels for an extended period of time could be useful in a wide variety of tasks.  One such task is simply monitoring local blood component traffic and composition. More complex tasks include serving as \invivo\ communication (\sect{communication network}) or navigational (\sect{navigation network}) nodes to assist other \invivo\ nanorobots operating in neighboring tissue, performing diagnostic functions (\sect{coordination}) such as searching the vicinity for microtumor masses having acoustic profiles distinguishable from healthy tissue, and performing therapeutic functions (\sect{therapy}) such as directed power transmission into tumor masses for inducing localized hyperthermia.

\begin{table}[ht]
\begin{center}
\begin{tabular}{ll}
parameter    &   value \\ \hline
ringset length	& $L = 10 \,\micron$ \\
vessel radius			& $\Router = 4 \,\micron$\\
inner radius of ringset	& $\Rinner = 3\,\micron$\\
\end{tabular}
\end{center}
\caption{\tbllabel{ringset geometry}Ringset parameters.}
\end{table}

\fig{ringset geometry} shows the geometry of the ringset we consider, located on the wall of a small vessel, with parameters given in \tbl{ringset geometry}. The ringset in this illustration consists of 200 individual robots, each of about $1\,\micron^3$ volume: twenty such robots form a ring around the inner wall of the vessel and the ringset consists of 10 such rings. 
For comparison with the pulsating sphere discussed above, the ringset is similar in size to the sphere with $5\,\micron$ radius.

The ringset is axially symmetric and we consider surface oscillations that also have this symmetry.  That is, all the robots in a ring coordinate to move their surfaces in the same way. This allows solving \eq{wave equation} numerically in a two-dimensional slice of the full three-dimensional domain using the finite element method~\cite{strang73}. Unlike the analytic solution for the sphere in \sect{sphere}, the numerical solution requires a finite domain which we specify as a sphere, of radius $R$, surrounding the ringset. We impose radiation boundary conditions for outgoing waves on this sphere to approximate the behavior of an unbounded domain.
To resolve the sound waves and forces on the ringset, we constrain the computational mesh size to be at most $0.1\,\micron$ on the ringset surface and at most $0.5\,\micron$ elsewhere. From \tbl{sound parameters}, this corresponds to at least 30 mesh elements per wavelength at $100\,\MHz$. This results in about 1300 mesh elements along the ringset boundary and $2\times10^5$ elements throughout the solution domain.

Robots in a ringset could actuate their surfaces in a variety of ways. One example is oscillating the outer surface next to the vessel wall as shown schematically in \fig{ringset movement}. To avoid discontinuity in the surface and improve convergence of the numerical solution, the range of movement is reduced smoothly to zero at the ends over a range of $0.1\,\micron$. Other motions include oscillating the inner (luminal) surface only, both inner and outer surfaces, or the surfaces at one or both ends of the ringset. Even within the context of axially symmetric oscillations, these options allow considerable variation, including different phases or magnitude of oscillation along the length of the ringset, i.e., parallel to the vessel axis.

\begin{figure}[htbp]
\begin{center}
\begin{tabular}{cc}
\includegraphics[width=3in]{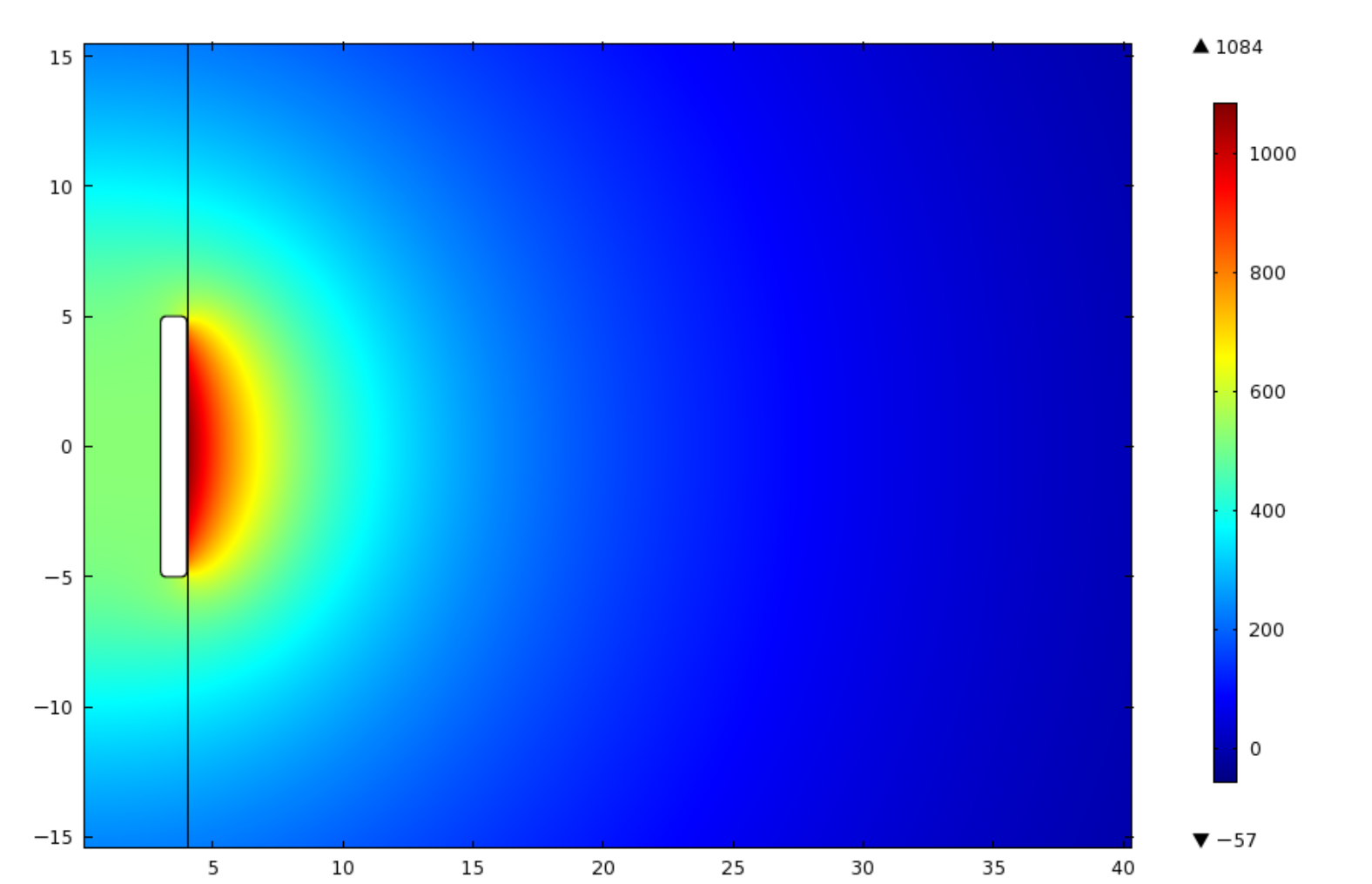}  & \includegraphics[width=2.5in]{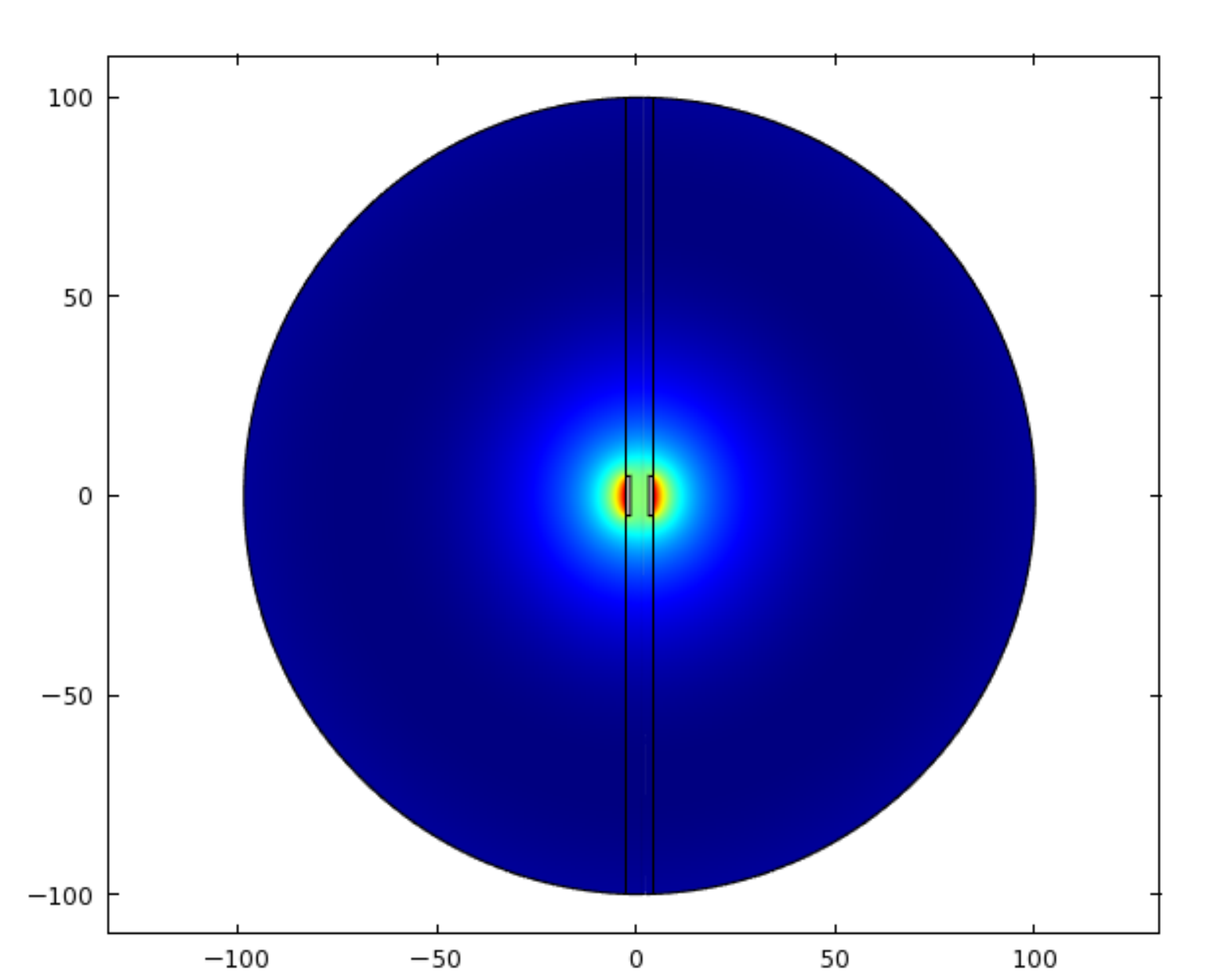} \\
\includegraphics[width=3in]{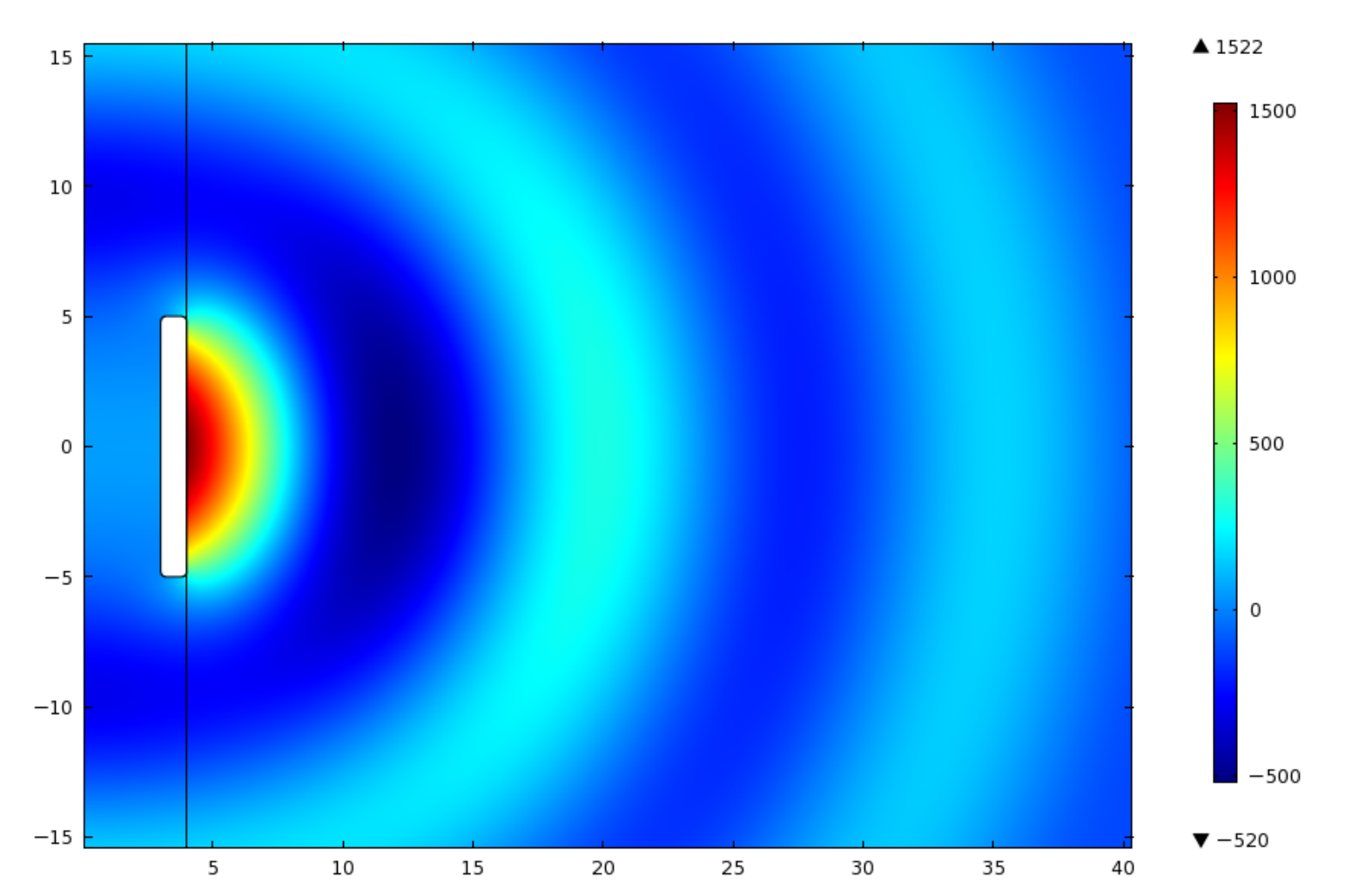}  & \includegraphics[width=2.5in]{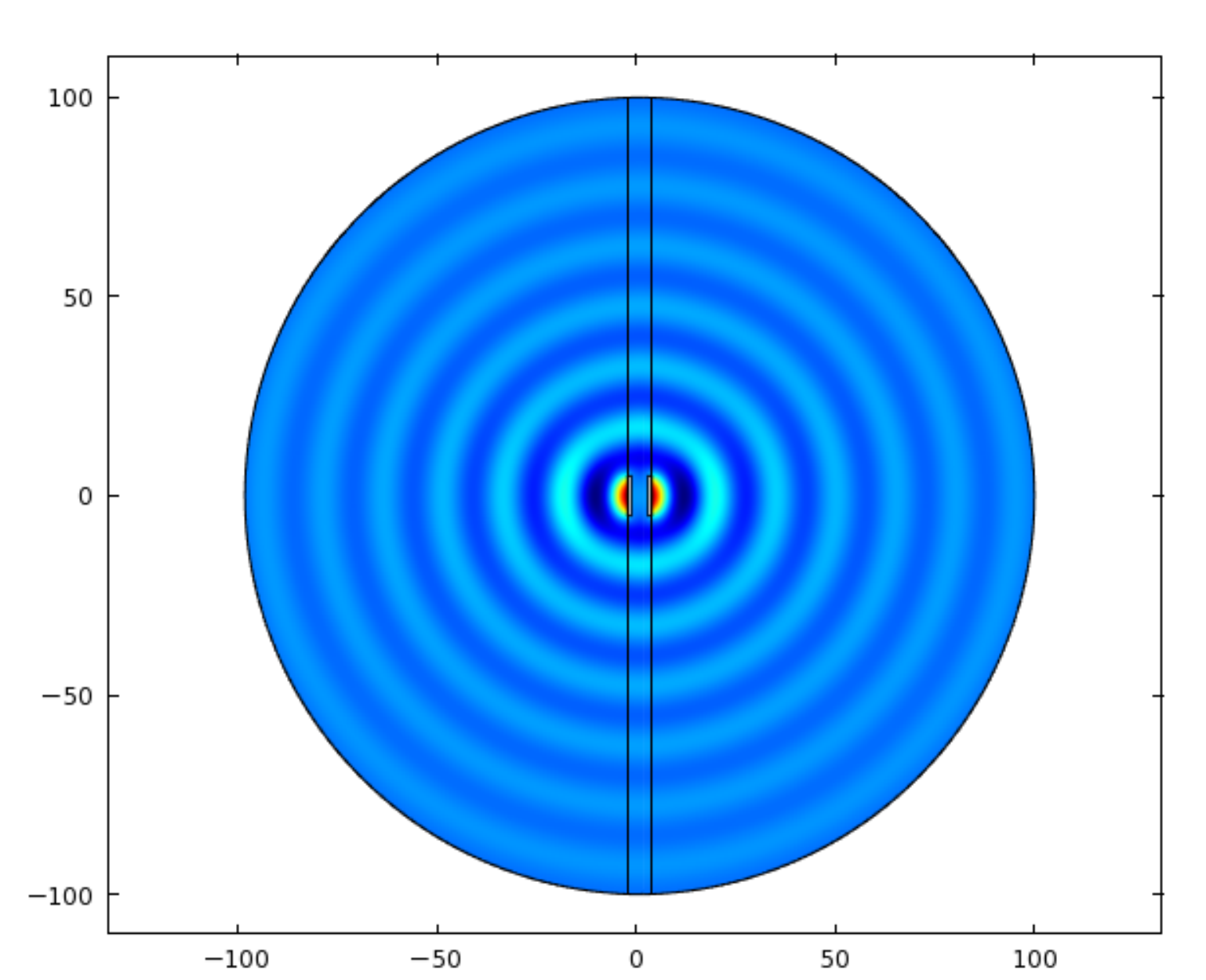} \\
\end{tabular}
\end{center}
\caption{Pressure variation for robot ringset for $10\,\MHz$ (top) and $100\,\MHz$ (bottom) at the times during the oscillation with the highest pressure, for low-attenuation tissue. Distances along both axes are in microns, and pressure is in $\Pascal$. The images on the left show the pressure field near the ringset, while those on the right show the full domain for the numerical solution. The vertical lines indicate the locations of the vessel walls, and the white rectangles indicate the robots next to the vessel wall.
}\figlabel{ringset pressure}
\end{figure}

\fig{ringset pressure} shows an example of the acoustic behavior near a ringset with oscillations on its outer surface.
Narrow vertical black lines indicate the locations of the vessel walls. Our numerical solution makes no use of these walls. This choice approximates the negligible effect of vessel walls for sound propagation. Furthermore, typical fluid motions in small vessels have little effect due to the high speed of sound compared to flow velocities, and so we neglect such motions in the numerical solution. Hence this plot is valid for ringsets located both in tissue and within a small blood vessel.
To compare scenarios, we pick the oscillation amplitude so that the time-averaged power, given in \eq{average power}, is $100\,\picoWatt$ in \tbl{ringset results}. The behavior for the ringset with oscillations on the outer surface is similar to that of a $5\,\micron$ sphere (\tbl{sphere results}). 
However, unlike the sphere, the oscillation of the ringset surface gives significant directionality to the beam at higher frequencies, as seen for the robot ringset in \fig{ringset power flux}.

\begin{table}[htb]
\begin{center}
\begin{tabular}{lcc}
frequency	 ($\MHz$)	& 10	& 100 \\
\hline
\multicolumn{3}{c}{water} \\
radiated power at $100\,\micron$ ($\picoWatt$) & $93$	&  $94$\\
average power flux at $100\,\micron$ ($\picoWatt/\micron^2$) & $7.4\times 10^{-4}$	& $7.5\times 10^{-4}$ \\
maximum pressure (\Pascal) 	& $1200$	& $1500$	\\
\hline \multicolumn{3}{c}{low-attenuation tissue} \\
radiated power at $100\,\micron$ ($\picoWatt$) & $76$	&  $90$\\
average power flux at $100\,\micron$ ($\picoWatt/\micron^2$) & $6.1\times 10^{-4}$	& $7.1\times 10^{-4}$ \\
maximum pressure (\Pascal) 	& $1100$	& $1500$	\\
\hline \multicolumn{3}{c}{high-attenuation tissue} \\
radiated power at $100\,\micron$ ($\picoWatt$) & $30$	&  $52$\\
average power flux at $100\,\micron$ ($\picoWatt/\micron^2$) & $2.3\times 10^{-4}$	& $4.1\times 10^{-4}$ \\
maximum pressure (\Pascal) 	& $680$	& $1500$	\\
\end{tabular}
\end{center}
\caption{\tbllabel{ringset results}Behavior of ringset with oscillations on its tissue-contacting outermost surface with various attenuations given by \eq{attenuation}.}
\end{table}

\begin{figure}[htb]
\centering  \includegraphics[width=\figwidth]{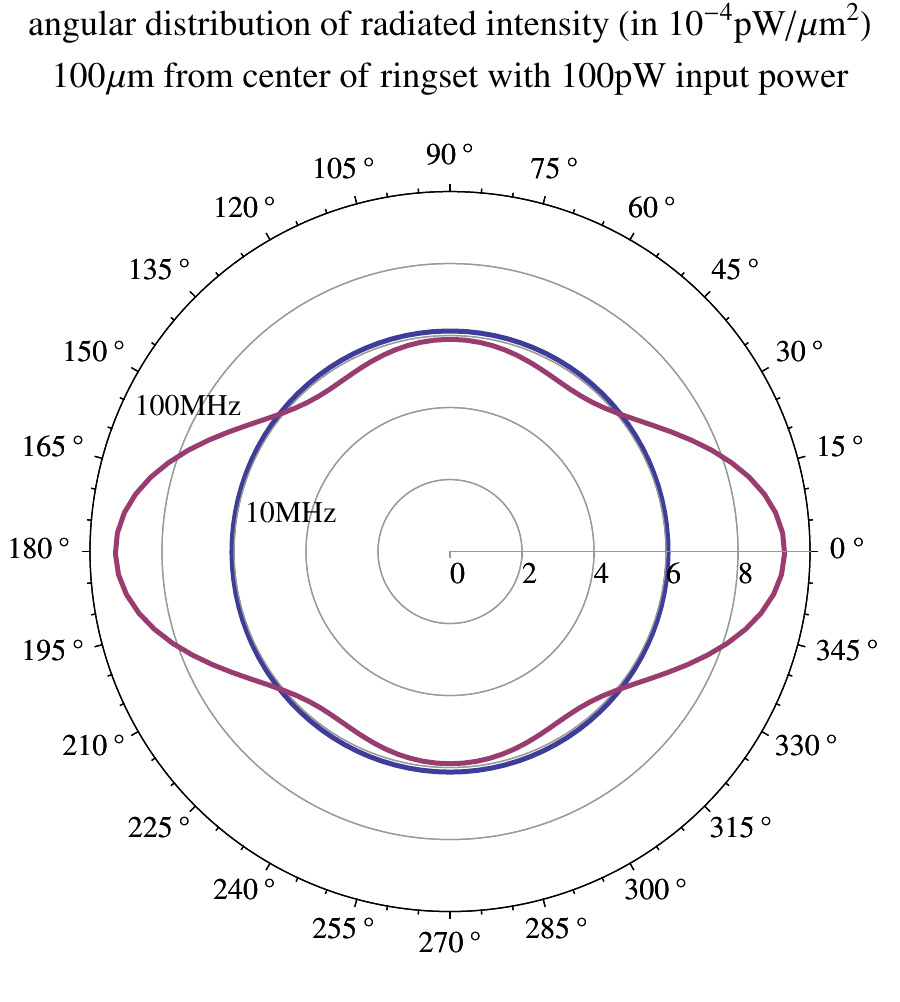} 
\caption{Power flux as a function of direction at $100\micron$ for the ringset at $10\,\MHz$ and $100\,\MHz$ in low-attenuation tissue. At $10\,\MHz$, the power flux of $\sim 6\times 10^{-4}\,\picoWatt/\micron^2$ is nearly uniform with respect to direction.  Time-averaged input power is $100\,\picoWatt$.
In this plot, the vessel axis is vertical, i.e., in directions indicated as $90^\circ$ and $270^\circ$.
}\figlabel{ringset power flux}
\end{figure}

\section{Directional Broadcasting}
\sectlabel{directional}

Communication between specific robots could benefit from directed beams of sound rather than uniform broadcast in all directions. The uniformly pulsating sphere discussed above produces sound uniformly in all directions. However, oscillations that are not spherically symmetric can give higher power flux in some directions than others, as indicated for the ringset in \fig{ringset power flux}. This raises the question of how tightly acoustic power can be directed.

A simple scenario to address this question is a vibrating disk embedded in a plane, where the disk moves a small distance periodically perpendicular to the plane in the manner of a simple cylindrical piston actuator~\cite{freitas99}. In this case, power flux is largest directly above the disk. At sufficiently high frequencies, the sound radiation pattern has certain directions with zero flux. A simple criterion for the width of this acoustic beam is  the angle of the first zero in the radiation pattern, which is given by~\cite{fetter80}
\begin{equation}\eqlabel{disk beam size}
\sin \theta = 1.22 \frac{\lambda}{d} 
\end{equation}
where $\lambda$ is the wavelength of the sound, $d$ is the diameter of the disk and the numerical factor is the first zero of the Bessel function $J_1$ divided by $\pi$. The main radiation lobe is within angles $-\theta$ to $\theta$ of the vertical axis of the disk. For long wavelengths compared to the size of the radiating object, the right-hand side is larger than one, so the equation has no solution. Instead, at long wavelengths the radiation pattern is fairly uniform in direction.
Another view of this directionality is the ratio of power flux in the direction with maximum flux to that of the average over all directions. This ratio, shown in \fig{disk directed power}, is approximately~\cite{fetter80}
\begin{equation}\eqlabel{disk flux ratio}
\frac{1}{2} \left( \frac{\pi d}{\lambda} \right)^2 
\end{equation}
for short wavelengths. At long wavelengths, the ratio approaches one, i.e., the radiation is nearly the same in all directions.

\begin{figure}[htb]
\centering \includegraphics[width=\figwidth]{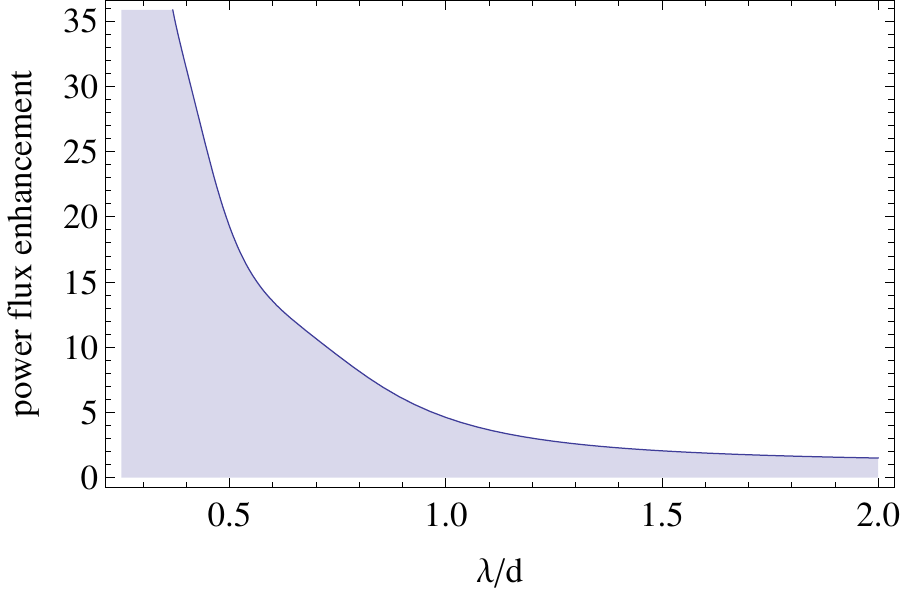}
\caption{Ratio of time-averaged power flux in the direction of maximum value to the average over all directions, for a vibrating disk in a plane as a function of the ratio of sound wavelength to disk diameter $\lambda/d$.}\figlabel{disk directed power}
\end{figure}

As an example, at $100\,\MHz$ the wavelength is $15\,\micron$ so a $10\,\micron$ disk is too small to produce a fully developed beam, i.e., where intensity goes to zero at one or more angles. That is, there is no angle $\theta$ satisfying \eq{disk beam size}. Nevertheless, the direction with maximum flux has about twice the average flux (corresponding to the value for $\lambda/d=1.5$ in \fig{disk directed power}). By contrast, a $100\,\micron$ disk significantly concentrates the radiated power, with maximum flux about 200 times larger than the average and \eq{disk beam size} gives $\theta = 11^\circ$ so the main beam extends over $2\theta = 22^\circ$.
Thus, concentrating power flux requires frequencies high enough that the sound wavelength is comparable to or smaller than the size of the oscillating surface.

\begin{figure}[htb]
\centering \includegraphics[width=\figwidth]{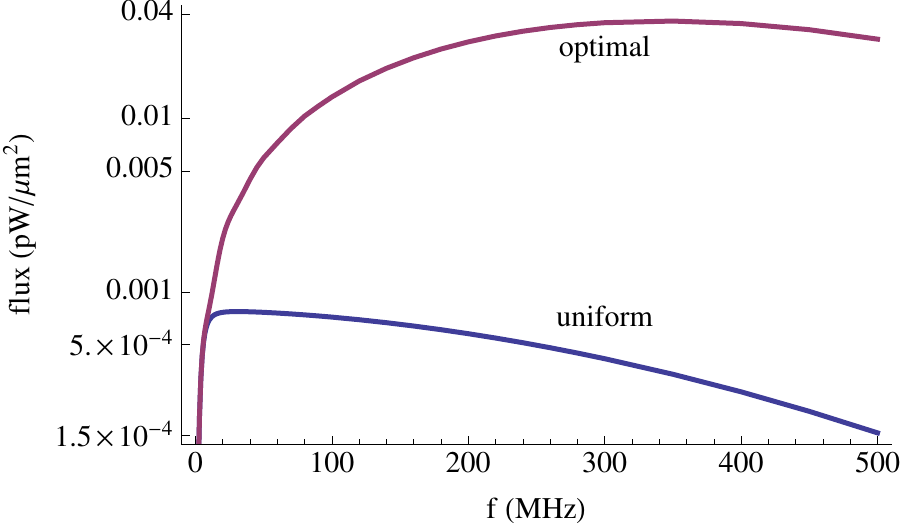}
\caption{Power flux for uniform and optimally directed oscillations (lower and upper curves, respectively) at $100\,\micron$ from a sphere with $5\,\micron$ radius in low-attenuation tissue. The sphere uses $100\,\picoWatt$ to generate the sound.}\figlabel{sphere directed power}
\end{figure}

Other geometries have similar behavior, though with somewhat different numerical factors, when oscillations are not spherically symmetric. Such oscillations can arise from the geometry, as with the uniform motion of a surface of the robot ringset or vibrating disk, or from actuators imposing nonuniform motions on a surface. In a nonuniform oscillation, different parts of the surface can have different amplitudes of motion as well as different phases, i.e., when one part of the surface is moving up another part is moving down. In the case of the sphere, arbitrary oscillation patterns are conveniently described as a sum of modes, each corresponding in shape to a spherical harmonic~\cite{abramowitz65,lebedev72}. The sound resulting from such modes is a generalization of the above discussion for uniform oscillations of a sphere~\cite{tsoi69}. In particular, this solution is useful for identifying how actuators should move the sphere's surface to give the maximum power flux in a desired direction when applying a given amount of power to produce the sound.

For example, \fig{sphere directed power} shows how well nonuniform surface oscillations can direct power in a specific direction for a vibrating sphere. For instance, around $100\,\MHz$, directional transmission increases the flux at the receiver by about a factor of 20 in low-attenuation tissue. The maximum directed flux is around $300\,\MHz$, with an enhancement by about a factor of 80 over uniform motion.
For high-attenuation tissue, the maximum directed flux occurs around $100\,\MHz$, with an enhancement of about a factor of 10 over uniform motion.
In accordance with the behavior seen with the vibrating disk, significant directional enhancement requires frequencies high enough for sound wavelength to be comparable to or smaller than the size of the sphere. \fig{sphere directed power} also shows that the frequency giving the largest flux in a specific direction is somewhat higher with nonuniform oscillations than with uniform motion. That is, the benefit of the shorter wavelengths with nonuniform oscillations more than compensates for the increasing attenuation at the higher frequencies. At the higher frequencies the optimal motions involve relatively high oscillation speeds at the surface and hence larger viscous power losses. 

As with the other discussions in this paper, this result only considers power expended against the fluid, not any power losses internal to the robot. The optimal motion at high frequencies involves changes in motion on the sphere surface over short distances and times, so would likely incur more internal power losses than uniform motion. Moreover, the motions require precise coordination in both space and time to produce the directed beam, especially at higher frequencies, thereby placing stronger requirements on the accuracy of the actuator controls than needed for uniform motion. This additional precision requirement leads to a greater sensitivity to noise within the robot.
\remove{These nonuniform motions involve ``bumps'' moving on the surface at substantial fractions of the speed of sound? For instance, a bump moving from one side of the sphere to the other over 1/2 an oscillation period, moves $\pi a \approx 15\,\micron$ in $0.5/f = 0.005\,\microsecond$ for $100\,\MHz$ sound; corresponding to $3000\,\meter/\second$. Just as with waves on water, this is not a speed at which material of the sphere surface is moving; rather it indicates how rapidly the oscillation wave is moving across the surface, and hence how rapidly actuators must coordinate their actions.}

The ringset described in \sect{ringset} can significantly concentrate sound in a particular direction by actuating its surface nonuniformly provided the frequency is high enough. \fig{ringset power flux delay} is one example. In this case, the outer surface of the ringset oscillates at $100\,\MHz$ with uniform amplitude (except for the smoothing at the ends indicated in \fig{ringset movement}) but with different phases along the direction parallel to the vessel axis. Specifically, in this example the phase of surface motion at distance $z$ along the vessel is $\exp(i \omega z/c)$ where $c$ is the speed of sound.
By contrast, at $10\,\MHz$ power flux is nearly uniform with respect to direction even with nonuniform oscillations.
The axial symmetry used in the numerical solutions for the ringset means directed beams can only be along the vessel axis. However, surface oscillation patterns without this symmetry can direct sound in other directions. Thus, with suitable robot geometry and preprogrammed acoustic emission patterns, the ringset can preferentially broadcast in specific directions away from the vessel provided the frequency is sufficiently high.

\begin{figure}[htb]
\centering  \includegraphics[width=\figwidth]{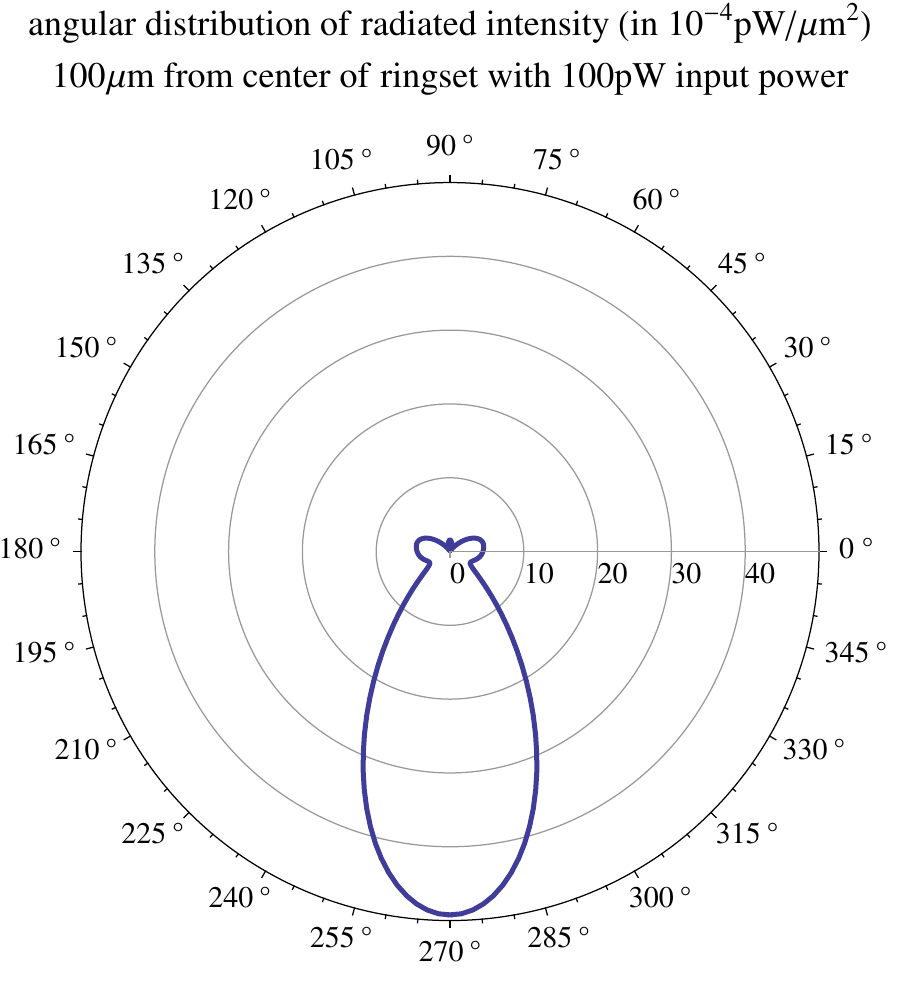} 
\caption{Power flux as a function of direction at $100\micron$ for the ringset with nonuniform surface oscillation at $100\,\MHz$ in low-attenuation tissue. Time-averaged input power is $100\,\picoWatt$.
In this plot, the vessel axis is vertical, i.e., in directions indicated as $90^\circ$ and $270^\circ$. Thus this motion of the outer surface radiates mainly in one direction along the vessel.
}\figlabel{ringset power flux delay}
\end{figure}

Thus significant directionality to the acoustic radiation requires wavelengths small compared to the size of the radiator. This can be achieved by using higher frequencies (which also give higher attenuation) or larger radiators, e.g., larger groups of robots. Larger radiators can consist of several small groups of robots spaced some distance apart along the vessel, or multiple ringsets either contiguous (effectively making one very long ringset) or spaced apart.
By accurately coordinating the time and amplitude of their surface oscillations, the group as a whole could act coherently as a larger radiator and thereby produce directed beams at lower frequencies than would be possible for one of the groups acting alone. Larger radiators could also consist of one or more groups of somewhat larger robots.
Thus, with a fixed number of robots, spacing them far apart gives more directionality (at a given frequency); though with the tradeoff that it gets harder to ensure they are accurately moving relative to each other. In particular, the surface motions involved in optimal sound direction correspond to robot actuators arranging for oscillation waves on the robot surface that move across the robot surface at a significant fraction of the speed of sound in the surrounding fluid. So widely separated robots would have to accurately know their distance to the other robots and have well-synchronized clocks or other control protocols~\cite{freitas99} to give the correct actuation pattern.

Directionality also applies to receiving the sound. Specifically, a robot could detect sound at several locations on its surface. Forming weighted combination of these inputs can arrange for highest sensitivity for sounds from a specific direction. Alternatively, if interference from other sound sources at the same frequency is an issue -- such as other communicating nanorobots -- a direction of minimum reception could be selected to reduce reception from those undesired sources~\cite{meyer02,argentieri06}.

\section{Biological Constraints on Ultrasound}
\sectlabel{safety}

Ultrasound has been safely applied to the human body for medical purposes for more than 60 years.  It is most often employed in diagnostic applications, such as organ imaging in cardiology and gastroenterology, blood vessel patency and flowmetry in cardiovascular and neurology applications, bladder status in urology, and fetal imaging in obstetrics and neonatology.  It also has more limited therapeutic applications including focused ultrasound surgery in lithotripsy, localized heating to treat cysts and tumors, procoagulation ($5\mbox{--}12\,\MHz$), blood-brain barrier penetration for drug delivery, cataract treatment using phacoemulsification, and tooth-cleaning in dentistry.  Soft-tissue diagnostic ultrasound generally uses frequencies in the $1\mbox{--}20 \,\MHz$ range, but specific frequency choices depend on the tissue examined, e.g., $1\mbox{--}5 \,\MHz$ for abdominal, head and heart scans, $5\mbox{--}20 \,\MHz$ for the eyes, and $30\mbox{--}70 \,\MHz$ for intravascular scanning.

Safety is an important consideration for medical use of nanorobots~\cite{freitas03}. The main factors in assessing biological safety are power, pressure, and heating constraints on safe robot operation. The remainder of this section discusses these constraints.

\subsection{Acoustic Nanorobot Power Constraints}
\sectlabel{safety:power}

No significant biological effects have been reliably observed in mammalian tissues exposed \invivo\ to unfocused $\sim$MHz medical ultrasound with intensities up to $1000\,\Watt/\meter^2$~\cite{AIUM78} or after continuous exposures exceeding one second in duration for total energy transfers up to $500\, \mbox{kJ}/\meter^2$ (vs.~$\sim 0.5 \,\mbox{kJ}/\meter^2$ for a comparable exposure to UV excimer laser light~\cite{gower96}). However, exposures of any duration  above $500,000\,\Watt/\meter^2$ may cause cavitation and other harmful effects in biological tissue.  For example, stable cavitation can occur when small pre-existing bubbles surrounded by water resonate in synchrony with the acoustic field, with the liquid acting as the oscillating mass and the gas serving as the compliant component. An air bubble of radius $r_{\mbox{\scriptsize bubble}}$ (meters) in water has a resonant frequency $f_{\mbox{\scriptsize res}} \approx  3 / r_{\mbox{\scriptsize bubble}}$ (Hz) up to $\sim$1 MHz~\cite{minnaert33}. Thus a $6\,\micron$ bubble resonates at about $600\,\kHz$.  Resonating microbubbles have been reported in therapeutic beams at power intensities as low as $6800\,\Watt/\meter^2$ at $750\,\kHz$~\cite{mcdicken91}.  Transient cavitation in water requires about $10^5\,\Watt/\meter^2$ at $30\,\kHz$ and $10^6\,\Watt/\meter^2$ at $1\,\MHz$~\cite{mcdicken91} but intensities less than around $10^4\,\Watt/\meter^2$ will not produce transient cavitation in any tissue~\cite{bushong91}.  Shock waves can most easily form in liquids having low attenuation such as urine in the bladder or amniotic fluid;  a $3\,\MHz$ $10^6\Pascal$ pulse shows a shock waveform after passing through $5\, \mbox{cm}$ of water~\cite{mcdicken91}.  Dissipation of acoustic vibrational energy can initially heat tissues by about $1\,\Kelvin/\mbox{minute}$ if applied at, say, $50,000\,\Watt/\meter^2 $ at $3\,\MHz$.  Continuous exposure to $2000\mbox{--}6000\,\Watt/\meter^2$ at $0.1\mbox{--}10\,\MHz$ raises human tissue temperature by $1\,\Kelvin$ at equilibrium, which is considered safe~\cite{bushong91}. At the opposite extreme, ultrasound intensities of around $2 \times 10^7\,\Watt/\meter^2$ at the point of action are used to cauterize liver tissue after surgery.

Since 1985 the U.S. Food and Drug Administration (FDA) has allowed ultrasound intensities up to $7300\,\Watt/\meter^2$ for cardiac use, $15,000\,\Watt/\meter^2$ for peripheral vessels, and $1800\,\Watt/\meter^2$ for fetal, abdominal, intraoperative, pediatric, cephalic, and small-organ (breast, thyroid, testes) imaging~\cite{mcdicken91}. The American Institute of Ultrasound in Medicine~\cite{AIUM92} allows intensities as high as $10,000\,\Watt/\meter^2$ for exposures to highly focused sound beams, which is about the highest that human volunteers can tolerate~\cite{nyborg85}.  Thus a conservative limit is $10^4\,\Watt/\meter^2$, or $10^4\,\picoWatt/\micron^2$, as the maximum safe acoustic power dissipation. This limit applies to the highest intensities involved in an application, which will generally be at the robot-tissue interface (in the absence of constructive interference effects as discussed in  \sect{therapy}).

Power significantly constrains the
robots~\cite{mallouk09,soong00}, especially for long-term
applications where robots may passively monitor for specific rare
conditions (e.g., injury or infection) and must respond rapidly when
those conditions occur. The robots could obtain energy
from their environment, such as converting externally
generated vibrations to electricity~\cite{wang07} or
chemical generators\remove{, e.g., a fuel cell using glucose and
oxygen in the bloodstream}~\cite{freitas99,hogg10}, likely providing about $10\picoWatt$ in steady-state per micron-sized robot. With a 10\% duty cycle and sufficient onboard energy storage, each robot can radiate at $100 \picoWatt$ (for 10\% of the time).  For a 1\% duty cycle, which is still quite plausible and even useful, the maximum power for transmissions is $1000 \picoWatt$ for a micron-sized robot, well within the $10^4\,\picoWatt/\micron^2$ safety limit for a radiator surface of area around $1\,\micron^2$. Other possible power sources include onboard batteries or stored chemical reagents, and \exvivo\ acoustic sources most effectively using sub-MHz frequencies~\cite{freitas99}, in all cases respecting  the $10^4\,\picoWatt/\micron^2$ safety limit.

The previous discussion gave results for spherical robots (\sect{sphere}) and ringsets (\sect{ringset}) emitting a total acoustic power of $100\,\picoWatt$. For small individual robots, e.g., a sphere with $0.5\,\micron$ radius, the available power could be smaller than this value. Conversely, for larger robots or groups of robots (e.g., ringsets), the total available power from the collective could be much larger. 
In the case of a 200-robot ringset, $100 \,\picoWatt$ total power requires only $0.5\,\picoWatt$ from each robot.  Such a ringset will generally have at least 10-20 times this much maximum continuous power available per robot~\cite{hogg10}, or $1000\mbox{--}2000\,\picoWatt$, and total power can be even larger for short bursts of communication or for groups having more robots (e.g., a ringset of length $100\,\micron$) Ð but none of these instances would exceed the $10^4\,\picoWatt/\micron^2$ safety limit.  Longer ringsets of robots that use ambient oxygen for power will experience more competition for available oxygen among downstream robots~\cite{hogg10}, so power will not scale linearly with the size of the ringset but will plateau when ringset length or demand are large enough to maximally deoxygenate the passing blood.

\subsection{Acoustic Nanorobot Pressure Constraints}

In the context of acoustic communication, the primary concern is the effect of pressure variation on tissue.  
Power is proportional to the square of the surface oscillation amplitude whereas pressure is proportional to the amplitude, hence a 100-fold increase in power corresponds to a 100-fold increase in heating but only a 10-fold larger pressure variation.  The maximum pressure variation occurs close to the radiating robots.  For instance, robots emitting $100\,\picoWatt$ of acoustic power will generate a maximum of about $1000\,\Pascal$ pressure variations at or near their oscillating surface. Pressure variation rises to about $10^4\,\Pascal$ for a $10,000\,\picoWatt$ ringset.

With respect to robots and ringsets in blood vessels, pressure damage might most proximately affect passing erythrocytes, but under normal conditions nearly 1\% of all red cells are destroyed every day, roughly 3 million cells/s in the whole human body.  Increasing this destruction rate to 1.01\%/day, 1.1\%/day, or perhaps even higher should pose minimal risk to safety unless the red blood cell destruction is greatly increased and extremely localized.  The red cell plasma membrane membranolytic limit is $\approx 3 \times 10^6\,\Pascal$~\cite{skalak73,freitas99} with a red cell rupture strength $10^6 \,\Pascal$~\cite{chien75}. Impact hemolysis or ``march hemoglobinuria'' may occur with repeated exposures to low-frequency overpressures as low as $10^5 \,\Pascal$~\cite{freitas03}. Thus red cell damage at anticipated pressure variation levels seems unlikely.  Static fluid pressures of $42 \,\Pascal$ can initiate major changes in the endothelial cells in the arteries~\cite{fry68} and $300 \,\Pascal$ osmotic pressure produces enough tension, $\approx 2 \times 10^{-3}\,\mbox{N}/\meter$, to rupture eukaryotic cell membranes~\cite{nelson08}, but to date all mechanical cell stimulation experiments have been conducted at low frequencies ($<100 \,\mbox{Hz}$)~\cite{freitas03}.  Given the relative safety of procedures associated with intravascular ultrasound~\cite{weintraub94,gorge95,wong96,batkoff96,hamm97,rassin00,nissen01} with its low complication rate using frequencies as high as $10\mbox{--}20 \,\MHz$~\cite{weintraub94,gorge95,wong96}, it seems improbable that MHz acoustic waves of the intensities that might be employed by medical nanorobots for communication will damage the endothelial vascular walls.  Interestingly, relatively high-intensity intravascular ultrasound has been used to dissolve occlusive platelet-rich thrombi safely and effectively in myocardial infarctions~\cite{hamm97} and in restenosed stents~\cite{rassin00}.

The high-frequency nonuniform motions discussed in \sect{directional} in connection with \fig{sphere directed power} give much higher pressure variation at the surface of a radiating $5\,\micron$ sphere than do uniform motions, potentially introducing  safety issues when using directed beams even though the total power use and average flux over all directions remains the same. Yet even increasing acoustic flux up to 20-fold in a preferred direction at $100 \,\MHz$, or 80-fold at $300 \,\MHz$, only modestly increases surface pressure variations by 5-fold or 9-fold, respectively.

\subsection{Acoustic Nanorobot Heating Constraints}

Local heating of tissue near the robots is a potential safety concern, particularly for tissues with higher viscosity wherein a significant fraction of the input power is dissipated near the robot rather than producing outgoing acoustic radiation. For instance, consider a sphere with radius $a=1\micron$ and $P=100\,\picoWatt$ input power. Suppose, as a worst case, that all this power is dissipated near the sphere, the sphere generates this power continuously for a long time, and heat removal is only via conduction (i.e., ignoring any additional heat removal by convection in nearby blood flow). The power density for such a sphere is:
\begin{displaymath}
\frac{P}{\frac{4}{3}\pi r^3} \approx 2 \times 10^7 \,\frac{\Watt}{\meter^3}
\end{displaymath}
This power density is large compared to open-throttle gasoline-powered automobile engines and active neutrophils ($\approx 2\times 10^4 \Watt/\meter^3$)~\cite{freitas99}, or even to the $1\mbox{--}2 \times 10^6\,\Watt/\meter^3$ of tetanic skeletal muscle cells~\cite{wilson72}, myosin muscle motors~\cite{spudich94}, and bacterial flagellar motors~\cite{jones91}.  This large power density suggests localized heating could be significant and thus demands further analysis.

Quantitatively evaluating the heating effect requires determining the temperature near the robot.
The temperature distribution satisfies a diffusion equation:
\begin{displaymath}
\frac{\partial T}{\partial t} = \frac{\kThermal}{\rho \,\heatCapacity} \nabla^2 T
\end{displaymath}
where $\kThermal$ is the thermal conductivity and $\heatCapacity$ is the heat capacity of the tissue surrounding the robot, both of which we assume are constant. With the spherical symmetry, the steady-state temperature at distance $r$ from the center of the sphere is
\begin{displaymath}
T = \Tbody + \frac{P}{4\pi r \kThermal}
\end{displaymath}
With the values from \tbl{tissue parameters}, the largest increase in temperature (which occurs at the surface of the sphere, $r=a$) is less than $10^{-4}\,\Kelvin$.
This slight heating in spite of high power density is also seen with aggregates of a few hundred such robots~\cite{hogg10}. Thus local tissue heating appears not to be a serious safety concern, even in the extreme case of all the input power dissipated by viscous forces in the immediate vicinity of the robot. This is due to the extremely rapid diffusion of heat at these microscopic distances.

\section{Applications}

Creating nanorobots able to broadcast ultrasound provides multiple useful capabilities. This section discusses applications to communication, navigation, sensing and therapy.

\subsection{Communication Capability}
\sectlabel{communication network}

The foregoing analysis suggests that individual micron-sized robots transmitting with $100\,\picoWatt$ produce an average flux $\Pflux$ at a $100\,\micron$ distance of a few times $10^{-4}\,\picoWatt/\micron^2$ over the range of attenuations found in tissue.  To relate this to a communication rate, consider a receiver with area $A=1\,\micron^2$ which receives signal power $\Psignal=\Pflux A \approx 10^{-4}\,\picoWatt$. Ambient sounds in the body are mainly at frequencies below $100\,\kHz$~\cite{freitas99}. Thus for the communication frequencies we consider ($10\,\MHz$ or higher), as long as simultaneously broadcasting robots are spaced far apart or use different frequencies, the noise is primarily thermal, with power $\Pnoise = \BoltzmannConstant T \, \Delta f$ where $\Delta f$ is the frequency bandwidth used for communication~\cite{nyquist28}. At body temperature $T = 310\, \Kelvin$, $\BoltzmannConstant T = 4\,\zeptoJoule$. With $\Delta f$ small compared to the carrier frequency $f$, the acoustic transmission properties (e.g., attenuation) are nearly uniform over the range of frequencies used for the signal, i.e., $f \pm \Delta f$, so we use the value of $\Pflux$ at the center frequency $f$ to determine communication rates.

The maximum communication rate with arbitrarily small error is the channel capacity. For thermal noise this rate is~\cite{shannon63}
\begin{equation}\eqlabel{channel capacity}
\Delta f \log_2 \left( 1+ \frac{\Psignal}{\Pnoise} \right)
\end{equation}
This rate increases with bandwidth $\Delta f$ and approaches the limiting value $\Psignal/(\ln(2) \BoltzmannConstant T)$ for large $\Delta f$. For instance, with $\Psignal = 10^{-4}\,\picoWatt$, this limit is $3.4\times 10^4\,\bps$. A bandwidth of $\Delta f=200\,\kHz$, giving $3.2\times 10^4\,\bps$, is large enough to get close to this limit, while remaining small compared to the carrier frequencies we consider (i.e., at least $10\,\MHz$). 

Achieving this communication rate with arbitrarily low error rate requires error-correcting codes and grouping bits into packets~\cite{shannon63}. This increases communication latency as well as the computational and memory requirements of the nanorobots. Alternatively, a communication protocol without error correction has  latency determined by the speed of sound (\tbl{tissue parameters}), e.g., about $0.07\,\microsecond$ for communication over $100\,\micron$. 
In that case, the signal-to-noise ratio determines the communication rate with given error rate. Specifically, we define the signal-to-noise ratio by the logarithm of the ratio of signal power to noise power:
\begin{displaymath}
\SNR = \ln \left( \frac{\Psignal}{ \Pnoise} \right)
\end{displaymath}
The corresponding value in decibels is $10\, \SNR/\ln 10 \approx 4.3\, \SNR$.
For example, at a signal-to-noise ratio $\SNR=2$ (i.e., about $9\,\decibel$), the energy reception threshold is $E \equiv \BoltzmannConstant T  \,e^\SNR \approx 30\,\zeptoJoule$ at body temperature $T = 310\, \Kelvin$~\cite{freitas99}.
A receiver with this threshold and $\Psignal = 10^{-4}\,\picoWatt$ would receive about $10^4\,\bps$ from the transmitter.

The communication rate at a given frequency and distance can be improved in several ways:  (1) increase total receiver surface (e.g., more receivers per robot, more robots in aggregate, or larger robots), (2) increase transmitter power (e.g., more energy-dense power supply, onboard energy storage, and more robots in aggregate or larger robots), (3) use a shorter duty cycle with burst transmissions, or (4) use directed beams if the frequency is high enough to produce directed beams (\fig{sphere directed power}) for the given robot (or aggregate) size.  For instance, receiver area could be increased by a factor of 1000 using a ringset, boosting data rates to $10^7\,\bps$.  Operating 1000 robots in a ringset as a single transmitter using the same per-robot transmitter power as before, similarly achieves a continuous $10\, \MHz$ bit rate.  Reducing duty cycle to 0.1\% effectively increases the available broadcast power per transmission by 1000.  More specifically, consider a $100\,\micron$ ringset with an external (vessel-wall-contacting) radius of $4\,\micron$ that has a potential receiver area of about $2500 \,\micron^2$.  From \tbl{ringset results}, if only 10\% of the external surface area of this ringset is devoted to receivers then it can receive signals broadcast at $100 \,\MHz$ from $100 \,\micron$ away by another ringset having total transmitter power of about $2000 \,\picoWatt$.  Since the ringset consists of about 2000 micron-size cubical nanorobots, only about $1\,\picoWatt$ per nanorobot (in the ringset) is required to make the transmission.

Another characterization of communication efficiency is the energy required to transmit a bit of information. For instance, the example described above for transmitting $10^4\,\bps$ across a $100\,\micron$ distance using $100\,\picoWatt$ corresponds to $10^{-14}\,\Joule/\bit$. In practice, the communication energy requirement per received bit also includes the energy involved in computation~\cite{feynman96} by the transmitter and receiver. 

Aggregated or larger robots have several advantages compared to isolated or smaller ones.  First, the larger size has more available power, either by sharing power among multiple individual robots from internal generation, or by larger onboard generators or more capacious storage for higher burst power.  Second, the larger radiating surface in either case reduces losses to viscosity, which is especially significant in tissues with larger attenuation, hence giving higher acoustic efficiency.  Third, the larger size of aggregates or of individual robots compared to the sound wavelength increases the ability of the robots to direct the acoustic radiation.

Acoustically-enabled nanorobots capable of communicating over $100 \,\micron$ ranges could also form \invivo\ communication networks that could transfer data across much larger distances than possible with direct transmission due to attenuation at high frequencies~\cite{stojanovic06}.  A large number of communicating ringsets positioned in capillaries throughout a tissue mass could form a packet switching network similar to the internet.  This network could have multiple uses:  (1) sharing data between physically separated ringsets, (2) allowing individual nanorobots to communicate with distant ringsets or distant individual nanorobots, by communicating with the ringset through which the nanorobot was passing and using the ringset network as a message relay system;  and (3) as a means to aggregate and transmit data to \exvivo\ destinations, via implanted receiver nodes having direct physical connection to an external modality.  

Given the frequency of operation, the available power, the receiver sensitivity, and the physical spacing lattice of the ringsets, standard network theory can be applied to calculate useful network parameters for specific nodal architectures such as the maximum bit rate between nodes, total network traffic capacity, message latency, robustness of traffic to node failures, and so forth. 
As an example, for a sphere with $5\,\micron$ radius in low-attenuation tissue using $100\,\picoWatt$ to produce a directed beam, as discussed in \sect{directional}, gives a maximum flux of $5\times 10^{-5}\,\picoWatt/\micron^2$ at a distance of $1000\,\micron$ with a frequency of $80\,\MHz$. For a receiver with area $1\,\micron^2$, this corresponds to a maximum communication rate of about $2\times10^4\,\bps$ from \eq{channel capacity}.
On the other hand, $100\,\picoWatt$ could also power a network of ten such robots spaced $100\,\micron$ apart and each using $10\,\picoWatt$. This network could relay the message across $1000\,\micron$ in a series of ten $100\,\micron$ steps. Each robot in this network would produce one-tenth the flux of a single $100\,\picoWatt$ directed beam at a distance of $100\,\micron$ shown in \fig{sphere directed power}. Thus each robot would produce a flux of $4\times 10^{-3}\,\picoWatt/\micron^2$ at its neighbor in the network using a frequency around $350\,\MHz$. Provided these robots use somewhat different frequencies to avoid interference with each other, from \eq{channel capacity} this power flux corresponds to a maximum communication rate of about $10^6\,\bps$. Thus using a network to relay the message in ten steps of $100\,\micron$ benefits from using higher frequencies, and hence narrower directed beams, than are useful for communication over the full $1000\,\micron$ distance due to the larger attenuation at high frequencies.

Additional evaluation of tradeoffs for network design are beyond the scope of this paper but should be addressed in future work.  Also deferred to future work are certain special cases such as networks in highly inhomogeneous high-attenuation tissues like bone, where robots must operate in fluids confined to small channels rather than within the solid bone itself.

\subsection{Navigation Networks}
\sectlabel{navigation network}

Directional beams could be used to efficiently implement a navigational network inside the body.  Navigational networks would allow robot position sensing and aggregation of body position information for export~\cite{freitas99}.  Navigation networks have somewhat different requirements from communication networks.  For example the former has greater required precision of localization in space and stability of position over time.  Directionality is an important consideration.  Since acoustic transmission efficiency is fairly high in blood and most soft tissue, directionality can be increased by using somewhat higher frequencies (e.g., $\approx 200 \,\MHz$) to obtain tighter beams.  Higher power in shorter bursts at low duty cycle would avoid biocompatibility issues if not too extreme: up to $10,000 \,\Pascal$ peak intensity should be tolerable (\sect{safety}).  The narrower the width of the acoustic radiation distribution, the easier it will be for ringsets in adjacent capillaries to detect a small relative movement, which translates into a minimum detectable positioning error between each ringset.  Each ringset could be beaming burst packets to some number of adjacent ringsets and making continuous slight adjustments to the beam-out angles based on feedback from communicant ringsets indicating slight drift from the maximum signal.  
A network of such ringsets can apply various algorithms~\cite{freitas99}
to further reduce overall positional measurement errors.
Given the access to power chemicals, signal molecules, and ready mobility, capillary stationkeeping ringsets might be a better architecture for the navigation and communication network than embedding the devices in tissue as previously proposed~\cite{freitas99}.

The numerical calculation for the ringset gives the received power flux $\Pflux(R,\theta)$ as a function of direction where $\theta=0$ corresponds to the direction of maximum flux. 
As described in \sect{communication network}, a receiver can detect a change in energy of $E \approx 30\,\zeptoJoule$ with signal-to-noise of $\mbox{SNR}=2$. Thus if a receiver with area $A$ integrates the signal for a time $\Delta t$, it can recognize a change in orientation $\Delta \theta$ with respect to the maximum of the transmitter beam determined by
\begin{displaymath}
\Pflux(R,0) - \Pflux(R,\Delta \theta) = \frac{E}{A\,\Delta t}
\end{displaymath}
The corresponding minimum detectable positional drift of the receiver relative to a lobe peak is $\Delta X = R \sin (\Delta \theta)$.  
As an example, for the situation shown in \fig{ringset power flux delay}, a $1\,\micron^2$  receiver integrating the signal from a ringset at distance $R=100\,\micron$ for $\Delta t = 1 \millisecond$ gives a positional error of $\Delta X \approx 5\,\micron$.
Uncertainty can be reduced by combining information from multiple transmitters.
Increasing the size of the ringset enhances directionality at any given frequency.

Localization also may be improved by having the beam slightly off center of the maximum as a tradeoff between somewhat lower power vs.~a larger gradient (since power flux changes more rapidly somewhat away from the maximum)~\cite{yovel10}. Thus emphasizing localization instead of communication rate could give somewhat different designs. Ringsets could switch between broadcast modes optimal for localization or communication depending on their communication traffic load and the relative importance of these two processes at a given time.

\subsection{Coordinated Sensing by Robot Aggregates}
\sectlabel{coordination}

Aggregates of robots in one location for an extended period of time
could be useful in a variety of tasks. For instance, they could
improve diagnosis by combining multiple measurements of
chemicals~\cite{hogg06b} to give precise temporal and spatial control of drug
release~\cite{freitas99,freitas06}. Using
chemical signals, the robots could affect behavior of nearby tissue
cells. For such communication, molecules on the robot's surface
could mimic existing signaling molecules to bind to receptors on
the cell surface~\cite{freitas99,freitas03}. Examples include
activating nerve cells~\cite{vu05} and initiating immune
response~\cite{freitas03}, which could in turn amplify the actions
of robots by recruiting cells to aid in the treatment. Such actions
would be a small-scale analog of robots affecting self-organized
behavior of groups of organisms~\cite{halloy07}.
Aggregates could also monitor processes that take place over long
periods of time, such as electrical activity (e.g., from nearby
nerve cells), thereby extending the capabilities of devices tethered to
nanowires introduced through the circulatory system~\cite{llinas05}.
In these cases, the robots will likely need to remain on station for
tens of minutes to a few hours or even longer.

Aggregated robots could be useful as
computation hubs, e.g., for evaluating patterns of chemicals
detected nearby or communicated to the aggregate by other robots
as they pass by in the fluid. For treatment, computations shared
among many robots give more reliable decisions of whether and where
to initiate treatment. Moreover, requiring confirmation from other
robots to initiate an activity reduces failures due to errors by a
single robot.

In addition to helping form static structures, short-range communication among neighboring robots can enable dynamic structures with coordinated behaviors over much longer distances than that of individual inter-robot communication.  For example, swarming behavior for a group can arise when individuals are only able to detect activities of their immediate neighbors, provided the noise in such detection is below a transition threshold~\cite{vicsek95,levine00}. If robot sensors are too noisy to determine neighbor actions to get below this threshold, communication among neighbors can improve local coordination by exchanging intended actions among the neighbors, thereby avoiding the noise involved with robots attempting to directly sense their neighbors' actions. The resulting swarm behaviors can extend across the whole group, even though neighbor interaction distance is only a small fraction of the group size. That is, simple rules to respond to neighboring robot activities determined via short-range communication can produce long-range coordination.
In the case of a dynamic structure in a field of mobile robots flowing through a capillary at $\approx 1 \,\millimeter/\second$ past a fixed ringset communication hub, the robots are within 100 microns of the hub for $0.2 \,second$, enough to exchange 2000 bits during passage or 80 bits/passage per nanorobot assuming 25 1-micron-radius robots present in a 200-micron-long capillary volume at a 1\% nanocrit.  This should be a sufficient data rate to implement simple swarm algorithms, which require only a few parameters on neighbors such as their speed and direction of travel~\cite{vicsek95}.  Relative motions of robots within a laminar fluid flow should typically occur at equal or lower velocity~\cite{freitas99}, allowing equal or higher data rates within a dynamic aggregate without reference to the static vessel wall.

One application for swarms is improved detection of chemical gradients. While eukaryotic cells can orient in chemical gradients with as little as 1\% variation across a cell length~\cite{zigmond77}, this is more difficult for the considerably smaller bacterial cells comparable in size to the nanorobots we consider.  While counting rotors deployed at opposite ends of a 1-micron robot should detect a 0.1\% concentration differential of small common molecules over a $0.1 \,\second$ measurement period~\cite{freitas99}, measuring still shallower differentials may require evaluating chemical gradients over time by motion rather than at opposite sides of the cell, depending on the magnitude of the gradient~\cite{dusenbery98}. In the case of robots, communicating over a distance significantly larger than individual robots could aid in detecting and responding to spatial gradients rapidly. This is particularly useful for robots in vessels attempting to detect chemical sources on the wall of the vessel.  Fluid flow will make the largest concentration somewhat downstream of the source so a robot detecting the chemical would then have to move upstream, against the fluid motion, to the source, thereby requiring significant power. On the other hand, communicating the detection acoustically could notify upstream members of the swarm of the detection, giving them time to move toward the vessel wall prior to reaching the source and avoiding the large power requirements of moving against the fluid flow~\cite{hogg06a}.

Besides allowing improved measurement of subtle chemical gradients, nanorobot acoustic capabilities could enable acoustic-based detection of blood clots and tumor masses by the ringsets (e.g., acoustic tomographic scanning) for diagnostic purposes.  Later, after corrective measures (\sect{therapy}) have been applied, the same tissue could be re-examined to verify and validate completion of the designated therapy, or to repeat a procedure in the event of therapeutic incompleteness.  Comparing tissue properties at multiple frequencies (e.g., absorption) can provide additional diagnostic information to distinguish normal tissue from some forms of diseased tissue~\cite{narayana83,parker84}.  With an established longer range communication network, pulse arrival times and received power levels would also provide useful information about the tissue that was transited by the sound waves.  Acoustic emissions from robot aggregates on vessel walls could be useful for short range sensing of passing cells and the monitoring of physical blood variables such as velocity, hematocrit and viscosity.  Acoustic frequencies as high as $\approx 1 \,\GHz$ might also be relevant for nanorobots performing noninvasive transcellular acoustic microscopy~\cite{freitas99}, revealing internal features and components of tissue cells to a fine level of detail (i.e., $<1$ micron).

\subsection{Therapy}
\sectlabel{therapy}

Acoustic broadcast capabilities could be useful for treatment by enabling coordinated activity among neighboring robots, including aiding the robots to form aggregates. Such aggregates could provide structural support, e.g., in rapid response to injured blood
vessels~\cite{freitas00}. Aggregates\remove{ could also jointly
manipulate biological structures based on surface chemical patterns
on cells. Robots aggregated at chemically identified sites} could
perform precise microsurgery at the scale of individual cells,
extending surgical capabilities of simpler nanoscale
devices~\cite{leary06}. Since biological processes often involve
activities at molecular, cell, tissue and organ levels, such
microsurgery could complement conventional surgery at larger scales.
For instance, a few millimeter-scale manipulators, built from
micromachine (MEMS) technology, and a population of microscopic
devices could act simultaneously at tissue and cellular size scales,
e.g., for diagnosis~\cite{hogg09d} or nerve
repair~\cite{sretavan05,hogg05}. Acoustic communication among these robots over distances of about $100\,\micron$ would aid such tasks.

As an application of creating acoustic fields with specific intensity patterns described in \sect{directional}, robots with excellent positioning and timing could form a 3D network of emitters arranged in a carefully-designed 3D geometry whose overlapping emission patterns cancel and reinforce to create a 3D pattern having several acoustic hot spots where power is concentrated with relatively low power elsewhere. These high-intensity regions could be centered on multiple microtumors to destroy them. As one example, if a hot spot intensity of $10^5 \,\picoWatt/\micron^2$ is sufficient to raise microtumor temperature by $2\,\Kelvin$ per minute (\sect{safety:power}) and thus destroy tumor tissue after some tens of minutes of exposure, then ultrasound beams from 10 independent emitters, each operating at the safe $10^4 \picoWatt/\micron^2$ intensity, that overlap only within the tumor tissue volume should be sufficient to destroy it.  High-intensity regions could also be centered on targeted ringsets to supply them with significantly elevated power, or on multiple individual stationkeeping nanorobots with whom communication at very high bit rates is required, or on robots serving as energy dampers (absorbers) positioned at high-intensity nodes in the 3D pattern. If extending through inhomogeneous tissue, accurate design of these acoustic fields would also require precise knowledge of the varying acoustic properties of the tissue, which these robots could probe at lower intensities and adjust their high-power broadcasts based on preliminary measurements, provided the robots have sufficient computational capability to determine these adjustments. Alternatively, by positioning sensor robots near the regions of interest the group of robots could operate a feedback control loop to adaptively tune the acoustic field as is currently done on larger scales~\cite{widrow76}.

When employing nanorobots for treatment, maintaining external control over treatment activities is an important safety requirement. The acoustic communication network described in \sect{communication network} allows using  external authorization to switch from (passive) sensing to (active) treatment modes. As discussed in this section, the same acoustic capabilities needed for communication could also be deployed for treatment.

\section{Conclusions and Future Work}
\sectlabel{conclusion}

This paper described acoustic communication capabilities for \invivo\ nanorobots. We found robots could communicate readily at rates of $10^4\,\bps$ over distances of 100 microns within the limits of available power and safety constraints. Higher bit rates could be achieved with methods such as aggregation or short bursts of higher power.

There are numerous directions for future work based on these results.

At any given frequency, a larger ringset can achieve more directionality but, unlike the sphere, the ringset geometry is too complicated to easily find optimal nonuniform motions.  Future computational studies could explore the relevant ringset parameter space (ringset length, geometry, frequency, etc.) to find the actuation pattern on the ringset surface giving optimal focus of sound in any given direction.
An additional useful computational study would be of multiple emitters, both static and moving. The resulting interference patterns would identify situations leading to blind spots or hot spots in the combined acoustic field.

Future work should consider the effects of noise.  Beyond pure thermal noise (\sect{communication network}), there could be other noise sources in the $10\mbox{--}100\,\MHz$ range.  While, presumably, there are none that occur naturally in the body,  additional noise sources could arise if many robots attempt communication in the same region, e.g., when robots concentrate in a small volume of interest for treatment.  However, robots have a range of frequencies available for use. An interesting question is the number of communication channels potentially available for a group of nearby robots.  In sensing applications, it may also be possible to detect periodic variation in noise via stochastic resonance~\cite{wiesenfeld95} (a situation where a moderate amount of noise can be helpful).  This would require long-term broadcasting and thus would not be power-efficient, but it could be useful over longer distances by allowing detection with low signal-to-noise ratios.

This study ignored the robots' internal structure and the mechanism for producing sound by oscillating their surfaces. A question for future study is evaluating such mechanisms, particularly to identify additional power consumption due to dissipation within the robot. Due to the high stiffness and low dissipation of materials proposed for nanorobot designs~\cite{freitas10} compared with tissue, such additional energy losses are likely to be relatively minor.

Another question involving internal properties is whether the nanorobot as a whole would resonate at the acoustic frequencies. 
As an estimate of relevant resonant frequencies,   
consider a piston with stiffness $k_s \sim 25 \,\mbox{N/m}$ as typical for nanorobotic nanomachinery 
and mass $m  \sim 4 \times 10^{-15}\,\kg$ for a cubic micron nanorobot, then the lowest resonant frequency is $f_{\mbox{\scriptsize res}} \sim 10 \,\MHz$. There could be higher frequency resonances as well. 
Since this is toward the lower end of the range of $10-100 \,\MHz$ acoustic 
frequencies we consider, it may be necessary to select operating frequencies to avoid resonances in the robot.
Larger nanorobots or aggregates will have more mass and hence a lower resonant frequency.
By contrast, soft tissue is much less stiff than nanorobots and thus should have much lower resonant frequencies. However, the effects of tissue resonance on behavior of nearby nanorobots should be studied further.

We considered behavior over distances of $100\,\micron$ or so in homogeneous tissue. An interesting extension is to robots operating near the boundary of significantly different types of tissue (e.g., within bones).
The analysis presented in this paper should also be extended to relevant biological fluids and tissues possessing significant viscoelastic properties.

While fabrication of nanorobots will require significant technological developments, computational studies, such as presented in this paper, can help us predict their likely capabilities and produce designs that can safely exploit these capabilities.


\end{document}